\documentclass[11pt]{article}

\PassOptionsToPackage{table,dvipsnames}{xcolor}
\usepackage[final]{acl}

\usepackage{times}
\usepackage{latexsym}
\usepackage{soul}
\usepackage[T1]{fontenc}
\usepackage[utf8]{inputenc}
\usepackage{amsthm, amssymb, amsmath}
\usepackage{microtype}
\usepackage{inconsolata}
\usepackage{graphicx}

\usepackage[table,dvipsnames]{xcolor} 
\usepackage{tcolorbox}
\usepackage{enumitem}
\usepackage{subfig}
\usepackage{array}
\usepackage{booktabs}
\usepackage{pdfpages}
\usepackage{multirow}
\usepackage{supertabular, longtable}
\usepackage{graphicx}
\usepackage{amsmath}
\newcommand{\mstd}[2]{\(\text{#1}_{\scriptscriptstyle \pm \text{#2}}\)}

\definecolor{lightblue}{HTML}{d9ecff}

\tcbuselibrary{skins,breakable}

\title{Universal Activation Verbalizer: A Unified Framework \\for Cross-Model Activation Explanation}

\author{
  Haiyan Zhao$^{1}$ \quad
  Zirui He$^{1}$ \quad
  Guanchu Wang$^{2}$ \quad
  Ali Payani$^{3}$ \\
  {\bf Yingcong Li$^{1}$ \quad
  Mengnan Du}$^{4}$\textsuperscript{†}\\
  $^{1}$New Jersey Institute of Technology \quad
  $^{2}$University of North Carolina at Charlotte \\
  $^{3}$Cisco Research \quad
  $^{4}$The Chinese University of Hong Kong, Shenzhen \\
  \texttt{\{hz54,zh296,yingcong.li\}@njit.edu} \quad \texttt{gwang16@charlotte.edu}\\
  \texttt{apayani@cisco.com} \quad \texttt{mengnandu@cuhk.edu.cn}\\
  \textsuperscript{†}Corresponding author
}

\begin{document}
\maketitle
\begin{abstract}
Activation verbalization explains hidden representations in natural language, but existing methods are mostly limited to self-explanation, where each model explains only its own activations.
We introduce {\underline{U}niversal \underline{A}ctivation \underline{V}erbalizer} (UAV),  a shared-decoder
framework that uses donor- and layer-specific adapters to explain
activations from heterogeneous donor models.
UAV learns a lightweight adapter that converts donor activations into soft tokens in decoder's embedding space, and further supports adapter-only transfer by reusing a frozen decoder-side LoRA while training only a new adapter for another donor.
Across classification, fact retrieval, and gist summarization, UAV remains competitive with strong self-explanation baselines while enabling cross-model verbalization across model families and scales.
Ablations show that decoder-side tuning mainly improves task behavior, whereas the adapter provides the activation-grounded factual and semantic evidence. Code and data are available at our \href{https://github.com/hy-zhao23/ActExp}{GitHub repository}.
\end{abstract}

\section{Introduction}\label{sec:intro}
Representation learning has been fundamental to the success of large language models (LLMs)~\cite{achiam2023gpt,team2023gemini,liu2024deepseek}. 
However, understanding what information is encoded in their internal activations remains a central challenge. 
Prior interpretability work has studied activations from several complementary perspectives. 
Probing methods test whether linguistic, syntactic, or semantic properties are decodable from hidden states~\cite{belinkov2022probing,alain2017understanding}, while patching-based methods inspect representations by transplanting activations into carefully designed decoding contexts~\cite{ghandeharioun2024patchscopes,chen2024selfie}. 
In parallel, sparse autoencoders and dictionary-learning methods decompose activations into interpretable features, offering a bottom-up view of model internals~\cite{shu2025survey,bricken2023monosemanticity,cunningham_sparse_2023}.

Recently, activation verbalization has emerged as a more direct approach: instead of predicting predefined labels or discovering individual features, a language model is trained to describe the information contained in an activation using natural language~\cite{pan2026latentqa,karvonen2025activation}. 
This makes activation explanations more flexible and human-readable.
We view activation verbalization as an open-ended white-box readout
rather than ordinary classification or question answering. The tasks
test what information can be recovered from a particular donor layer:
the verbalizer receives the question and activation-derived soft tokens,
but not the source text. Such readouts may support auditing of concealed
information~\citep{karvonen2025activation}. 
However, existing training-based verbalizers are largely limited to \emph{self-explanation}, where a model is trained to explain only its own activations. 
This restriction makes them difficult to reuse across different model families, architectures, or scales, and can be inefficient when analyzing large donor models.

We address this limitation with \textsc{Universal Activation Verbalizer} (UAV), which explains activations from heterogeneous donor models with a shared decoder. Here, ``universal'' refers to decoder reuse across donors.
UAV maps donor activations into decoder-readable soft tokens through a trainable adapter, for which we study both MLP-style projection~\cite{liu2023visual} and Q-Former-style attention~\cite{li2023blip}.
The model is trained with activation-to-text alignment followed by explanation-oriented instruction tuning.
To improve transferability, we introduce adapter-only transfer, which freezes a decoder-side LoRA~\cite{hu2022lora} learned from one donor and trains only a new adapter for another. By training on cached activations without retaining the donor model in GPU memory, a 4B UAV decoder reduces the Stage-1 and Stage-2 costs from
7.1 and 26.5 to 2.7 and 11.6 GPU-hours per epoch, respectively, compared
with 12B self-decoding, with a validation-loss difference of 0.059 (See Appendix~\ref{app:efficiency}).

\begin{figure*}[t]
    \centering
    \includegraphics[width=\textwidth]{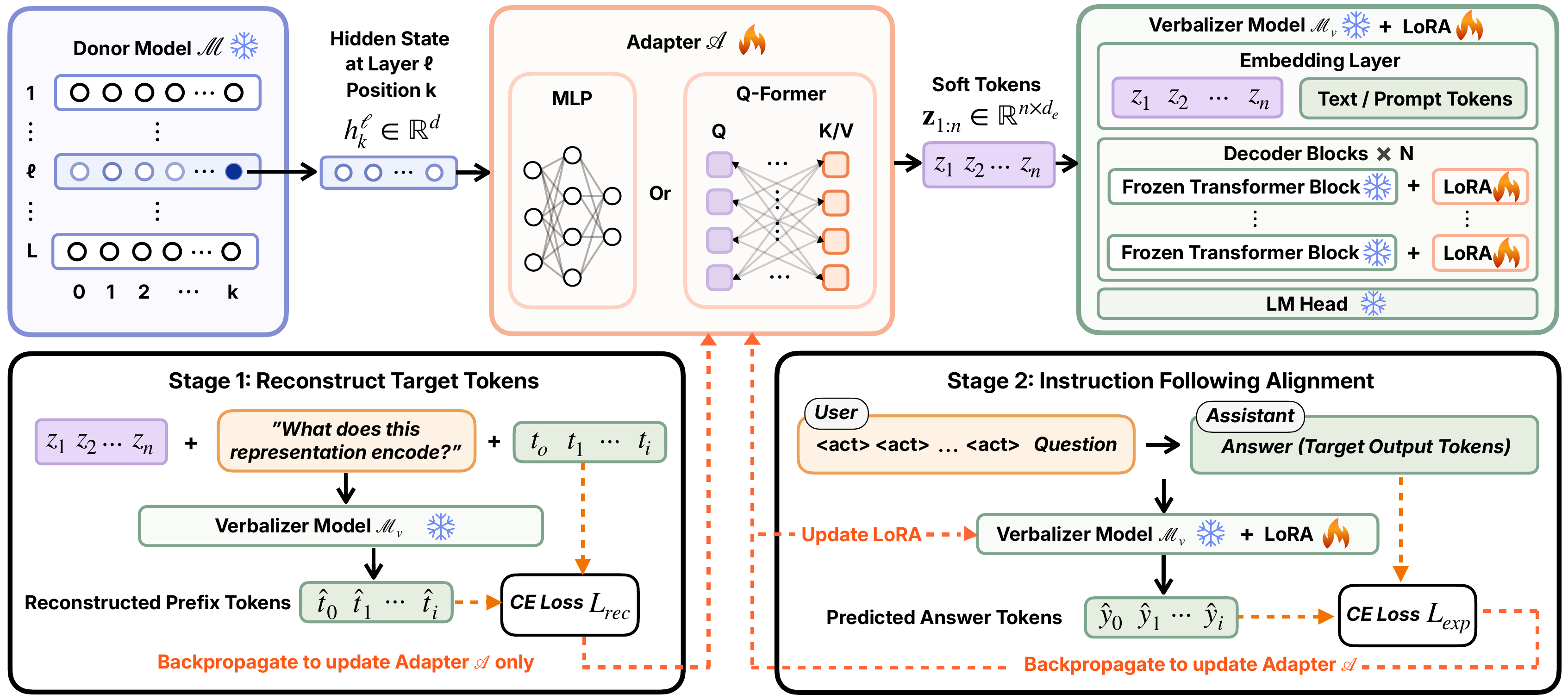}
    \caption{Overview of Universal Activation Verbalizer (UAV).}
    \label{fig:framework}
\end{figure*}

We evaluate UAV on generative activation-grounded QA across classification, fact retrieval, and gist summarization tasks.
UAV remains competitive with strong self-explanation baselines while enabling cross-model verbalization across Llama, Gemma, Yi, and Qwen-family donor models, with larger decoders generally improving performance.
Our analyses further show that decoder-side tuning improves task-level behavior, while the adapter is essential for recovering input-specific factual and semantic information from activations.

Our contributions are threefold:
(1) we propose UAV, a unified adapter-decoder framework for cross-model activation verbalization;
(2) we apply decoder-side LoRA reuse and donor-specific adapter training to cross-model activation verbalization;
and (3) we provide experiments and ablations showing that UAV supports activation explanation across model families and scales while remaining competitive with self-explanation methods.

\section{Related Work}\label{sec:related-work}

Prior works on explaining model activations can be categorized into bottom-up and top-down methods. Bottom-up approaches usually study what information is encoded in activations by defining or discovering interpretable features. For example, probing methods train lightweight classifiers on frozen representations to test whether linguistic, syntactic, or semantic properties are linearly decodable from activations~\cite{belinkov2022probing,belinkov2017neural,hewitt2019structural,rimsky-etal-2024-steering, conneau2018senteval, tenney2018what}. Meanwhile, sparse autoencoders (SAEs) perform unsupervised feature discovery by decomposing activations into sparsely-activated monosemantic latent features, which may correspond to concepts at different levels~\cite{huben2024sparse,bricken2023monosemanticity,cunningham_sparse_2023,achiam2023gpt}. However, this line of research is limited in directly providing human-interpretable explanations to activations. Probes query predefined properties, while SAEs decompose activations
into latent features. Activation verbalization instead provides an
open-ended natural-language readout of recoverable information
\citep{karvonen2025activation}.
UAV extends this interface across donor models using a shared external
decoder. It should be interpreted as a representation readout rather
than a causal or mechanistic explanation.

In contrast, top-down approaches aim to explain activations directly. Usually, LLMs are employed to verbalize the explanations toward target activations. In training-free paradigm, Patchscopes~\cite{ghandeharioun2024patchscopes} and SelfIE~\cite{chen2024selfie} sought to patch target activations directly into prompts, and utilize LLM to verbalize attributes and semantic information within. Alternatively, LatentQA~\cite{pan2026latentqa} and AO~\cite{karvonen2025activation} finetune LLMs to read their own activations from various layers by activation injection. But aforementioned approaches are incapable of cross-model explanations, which make them less generalizable and inefficient.

Lightweight adapters have been widely used to align heterogeneous representation spaces. In vision-language models, MLP projectors and Q-Former architectures map visual representations into language-model-compatible embeddings4~\citep{liu2023visual,li2023blip}. Adapter-based transfer and composition have also been explored in MAD-X~\citep{pfeiffer2020mad} and AdapterFusion~\citep{pfeiffer2021adapterfusion}. UAV builds on these ideas for cross-model activation verbalization: the adapter aligns donor activations with a shared decoder, while adapter-only transfer reuses a frozen decoder-side LoRA across donors.

\section{UAV Framework}\label{sec:uav-frame}

We introduce {UAV} framework that utilizes a trainable adapter to align donor activation space with the decoder model's embedding space~(See Figure~\ref{fig:framework}). In this work, we use representations and activations interchangeably.

\subsection{Problem Statement}
We study transformer-based LLMs whose last token representations are generally taken as an integration of information from all previous tokens. Give an input text whose input token sequence is $\langle\boldsymbol{t}_0, \ldots, \boldsymbol{t}_k\rangle$, we define activation at position $i$ as $\boldsymbol{h}_i$ that encoding all previous $i$ tokens. Our goal is to train a verbalizer $\mathcal{M}_v$ to answer questions related to a specific activation $\boldsymbol{h}_i^\ell$ from the $\ell$-th layer of donor model $\mathcal{M}$, where two models can be different. These questions can be trivial knowledge within the input sequence or general questions about gist of input text.

\subsection{Architecture Design}\label{sec:design}
To improve cross-model alignment, our framework consists of two stages:
(1) activation-to-text alignment and (2) instruction-following alignment.
In the first stage, we align the donor model's activation space with the
embedding space of the decoder model through input inversion tasks. This stage
serves as a warm-up that enables the adapter to translate activations into a
representation format that the decoder can effectively consume. In the second
stage, we further fine-tune the model so that it can follow explanation-oriented
instructions and generate natural-language explanations.

\paragraph{Activation-to-Text Alignment.} We freeze the base decoder model and fine-tune only the adapter on input inversion tasks. Given an input token sequence
$\langle\boldsymbol{t}_0, \ldots, \boldsymbol{t}_k\rangle$ and the corresponding donor
activation $\boldsymbol{h}_i$ at position $i$ $(0 \leq i \leq k)$, the goal of
input inversion is to reconstruct the prefix
$\langle\boldsymbol{t}_0, \ldots, \boldsymbol{t}_i\rangle$ from $\boldsymbol{h}_i$ under
teacher forcing. The formal training objective is provided in
Appendix~\ref{app:teacher-force}.

To perform this activation-to-text mapping efficiently, we compare two adapter
architectures: an MLP-based adapter~\cite{liu2023visual} and an attention-based
adapter following the Q-Former design~\cite{li2023blip}. The detailed designs
of both adapters are provided in Appendix~\ref{app:adapter-design}. As shown in
Figure~\ref{fig:framework}, the adapter bridges the donor activation space and
the decoder embedding space by mapping each donor activation $\boldsymbol{h}_i$
into $n$ continuous soft tokens
$\langle\boldsymbol{z}_1, \ldots, \boldsymbol{z}_n\rangle$. These soft tokens are prepended to the textual prompt tokens and injected at the embedding layer, serving as an activation-conditioned prefix that guides the
frozen decoder to reconstruct the  input prefix. We train the adapter
in a layer-wise manner, optimizing a separate adapter for activations from each donor layer.

\paragraph{Adapter-Decoder Framework.} In this stage, we continue training the adapter initialized from the activation-to-text
alignment stage, together with the decoder-side trainable parameters, to answer
questions about the input sequence. As shown in Figure~\ref{fig:framework}, the
second stage replaces the reconstruction objective with instruction-following
queries that target information contained in the input text. These queries cover
both fine-grained attributes, such as ``\textit{What was this person's profession?}'',
and topic-level understanding, such as ``\textit{What is the gist of this text?}''.
By jointly training the adapter and the decoder, this stage further
refines the activation-to-text alignment while improving the model's ability to
follow explanation-oriented instructions.

\begin{figure}[t]
    \centering
\includegraphics[width=0.75\columnwidth]{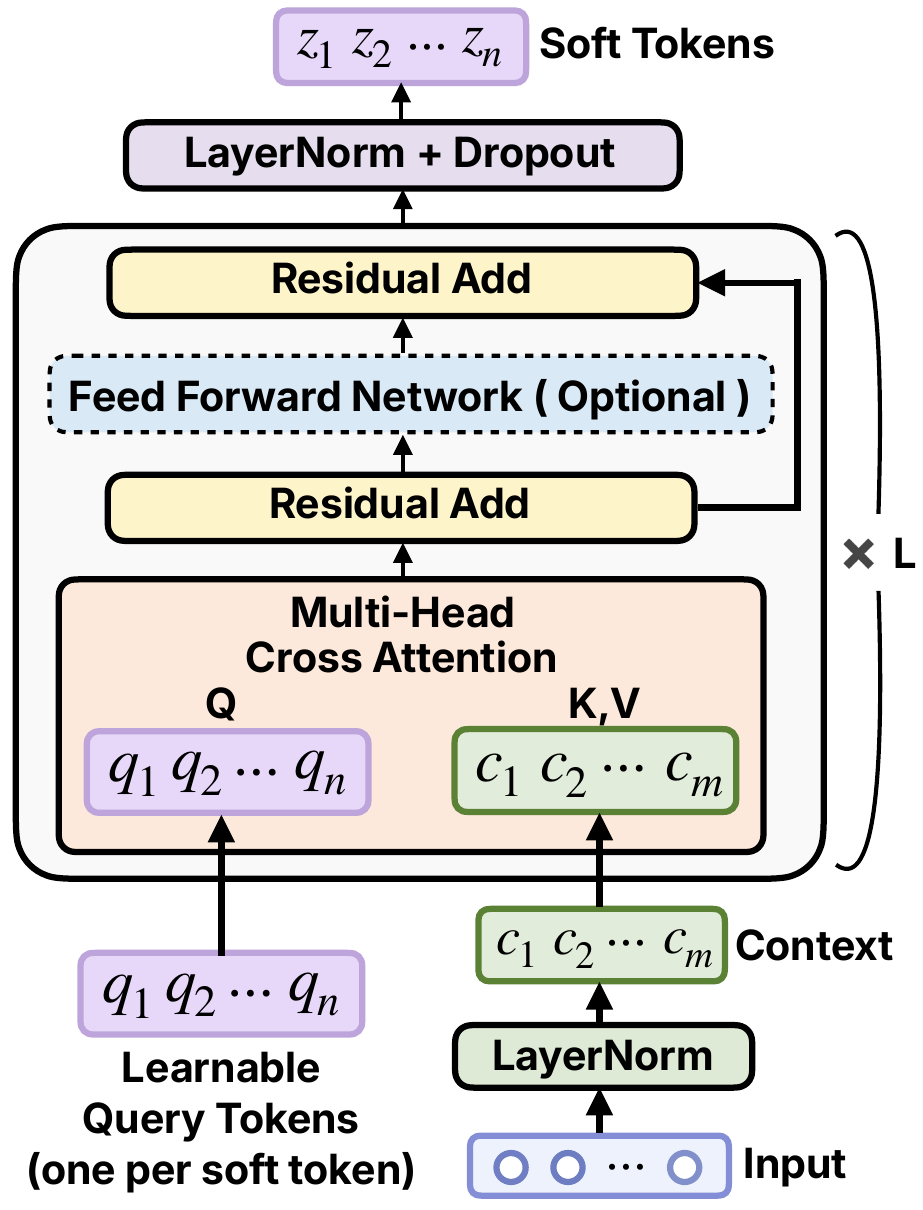}
\caption{Q-Former Architecture}
\label{fig:qformer_adapter}
\end{figure}

\subsection{Two Training Strategies}\label{sec:training}
UAV supports two adaptation strategies: full two-stage adaptation and adapter-only transfer (AOT). These strategies describe how the adapter and decoder-side LoRA are trained.

\paragraph{Full Two-Stage Adaptation.} In the first strategy, we use a common decoder backbone to explain activations
from different donor models, while performing the full two-stage training
procedure separately for each donor, as shown in Figure~\ref{fig:framework}.
Specifically, for each donor model, we first train a donor-specific adapter for
activation-to-text alignment with the frozen decoder. We then jointly fine-tune
the adapter and decoder-side LoRA parameters for explanation-oriented instruction
following. This strategy evaluates whether a shared decoder backbone can be
specialized to different donor models through donor-specific adapter and LoRA
training.

\paragraph{Adapter-Only Transfer.}In the second strategy, we test whether the instruction-following ability learned from one donor can transfer to others.
We first perform full two-stage adaptation on a source donor, then freeze the resulting decoder-side LoRA.
For each new donor, we train only a new adapter from scratch while keeping both the decoder backbone and LoRA fixed. This evaluates whether adapter-only training can align unseen donor activations to an already adapted decoder space.

\section{Experiments}\label{sec:experiments}
In this section, we first study the adapter design used for activation-to-text alignment.
We then evaluate UAV along four dimensions: effectiveness against prior methods, generality across donor models and decoder sizes, contributions of each training component, and sensitivity to activation layer, data scale, and activation-to-text warmup.
\subsection{Experimental Setup}\label{sec:exp-setup}
\paragraph{Models.}
We use Gemma-3-4B/12B~\cite{gemma-4b,gemma-12b}, Llama-3.1-8B~\cite{llama3-1-8b},
Yi-1.5-34B~\cite{yi-34b}, and Qwen3-4B~\cite{qwen-4b} as activation donors.
For decoder-side studies, we mainly use Qwen-family decoders, including
Qwen3-0.6B/4B/8B/32B~\cite{qwen-06b,qwen-4b,qwen-8b,qwen-32b} and
Qwen2.5-72B~\cite{qwen-72b}. 
We use \textit{self-explanation} as a broad term for methods that decode a model's own activations, including AO and LatentQA. Within UAV, we refer to a donor–decoder matched configuration as \textit{self-decoding} and a donor–decoder mismatched configuration as \textit{cross-decoding}. Throughout the paper, self-versus-cross describes the donor–decoder pairing, whereas Full-versus-AOT describes the adaptation strategy.

\paragraph{Datasets.}
We use generative question answering as the downstream task. 
Our training corpus is a 17-source mixture covering factual, scientific, topical, affective, classification-style, and conversational inputs, including Wikipedia entity snippets, peS2o scientific abstracts~\cite{peS2o}, AG News~\cite{zhang2015character}, SST-2~\cite{socher-etal-2013-recursive}, datasets reused from Activation Oracle~\cite{karvonen2025activation}, LMSYS-Chat-1M user turns~\cite{zheng2024lmsys}, and dair-ai/emotion tweets~\cite{saravia-etal-2018-carer}. 
All examples are tokenized with the Qwen3-4B tokenizer and capped at 64 tokens to match the activation extraction window. 
The resulting corpus contains 469K Stage~1 samples and 950K Stage~2 QA pairs, with full composition and preprocessing details in Appendix~\ref{app:datasets}. Classification and QA are used as readout tasks to measure what information is recoverable from donor activations, rather than as ordinary downstream prediction tasks.
\paragraph{Baselines.} We compare with both training-free and training-based activation verbalization methods. PatchScopes~\cite{ghandeharioun2024patchscopes} and
SelfIE~\cite{chen2024selfie} are representative training-free approaches, while
LatentQA~\cite{pan2026latentqa} and Activation Oracle (AO)~\cite{karvonen2025activation}
are latest training-based methods. We evaluate these baselines in the same
generative QA setting, using reconstruction- and QA-oriented
metrics to assess how well each method recovers information encoded in
activations.

\paragraph{Metrics.}
Since the questions are open-ended, we evaluate generated answers from
complementary perspectives, using both token- or character-level overlap metrics
and semantic similarity metrics. Specifically, ROUGE-L, chrF++, and token-level F1 measure sequence-level, character-level, and lexical overlap with the reference, respectively, while BERTScore captures semantic similarity using contextualized embeddings. Higher scores indicate stronger agreement between the generated answer and the reference text. Detailed definitions of all metrics are provided in Appendix~\ref{app:metrics}. The standard deviations reported in the main tables reflect variation across individual evaluation examples rather than across random seeds. We separately evaluate training stability using five Stage-2 runs under the 500K-QA setting, obtaining a validation loss of $1.8428 \pm 0.0012$ and a perplexity of $6.314\pm 0.007$. Full per-run results are provided in Appendix~\ref{app:seed}. We additionally evaluate source-grounded topic relevance and factual consistency using two LLM judges (See Appendix~\ref{app:llm_judge}). Paired-bootstrap 95\% confidence intervals support UAV's gains over AO
and LatentQA while indicating largely comparable self- and
cross-decoding performance; full results are provided in
Appendix~\ref{app:significance}.

\begin{table}[t]
\centering
\setlength{\tabcolsep}{2pt}
\caption{
Adapter architecture exploration.
We organize the ablations into two adapter families: MLP-style adapters and Q-Former adapters.
For MLP adapters, we vary the bottleneck size and the number of projected soft tokens.
For Q-Former adapters, configurations are denoted by $(L, C, r/\alpha)$.
}
\label{tab:adapter_arch}
\scalebox{0.64}{
\begin{tabular}{llccccc}
\toprule
\textbf{ID} & \textbf{Configuration}
& \textbf{Ada.} & \textbf{LoRA}
& \textbf{Val $\downarrow$} 
& \textbf{ROUGE-L $\uparrow$} 
& \textbf{BERTS $\uparrow$} \\
\midrule
\multicolumn{7}{l}{\textbf{MLP-style adapters}} \\
\midrule
\multicolumn{7}{l}{\textit{Bottleneck size}} \\
\textbf{ID} & \textbf{Bottleneck}
& \textbf{Ada.} & \textbf{LoRA}
& \textbf{Val $\downarrow$} 
& \textbf{Tok-F1 $\uparrow$} 
& \textbf{BERTS $\uparrow$} \\
MLP-B1 & 1024 & 195M & 132.1M & 1.6798 & 0.258 & 0.345 \\
MLP-B2 & 2048 & 367M & 132.1M & 1.6818 & 0.259 & 0.352 \\
MLP-B3 & 3072 & 540M & 132.1M & 1.6880 & 0.260 & 0.35 \\
\rowcolor{lightblue}
MLP-B4 & N/A  & 860M & 132.1M & \textbf{1.6738} & \textbf{0.296} & {0.353} \\

\midrule
\multicolumn{7}{l}{\textit{Number of projected soft tokens}} \\
\textbf{ID} & \textbf{$n$}
& \textbf{Ada.} & \textbf{LoRA}
& \textbf{Val $\downarrow$} 
& \textbf{Tok-F1 $\uparrow$} 
& \textbf{BERTS $\uparrow$} \\
MLP-T1 & 8   & 99.7M  & 132.1M & 1.6833 & 0.265 & \textbf{0.354} \\
MLP-T2 & 16  & 162.6M & 132.1M & 1.6921 & 0.259 & 0.352 \\
MLP-T3 & 64  & 540M   & 132.1M & 1.6880 & 0.260 & 0.350 \\
MLP-T4 & 128 & 1044M  & 132.1M & 1.7407 & 0.250 & 0.341 \\

\midrule
\multicolumn{7}{l}{\textbf{Q-Former adapters}} \\
\midrule
\textbf{ID} & \textbf{$(L, C, r/\alpha)$}
& \textbf{Ada.} & \textbf{LoRA}
& \textbf{Val $\downarrow$} 
& \textbf{Tok-F1 $\uparrow$} 
& \textbf{BERTS $\uparrow$} \\
QF-Base & $(1, 8, 16/32)$  & 162.8M & 33.0M  & 1.6818 & 0.267 & 0.351 \\
QF-Deep & $(4, 8, 16/32)$  & 398.8M & 33.0M  & 1.6863 & 0.272 & 0.352 \\
QF-Wide & $(4, 16, 16/32)$ & 482.7M & 33.0M  & 1.6755 & 0.278 & 0.357 \\
\rowcolor{lightblue}
QF-LoRA & $(2, 8, 64/128)$ & 241.4M & 132.1M & \textbf{1.6388} & \textbf{0.284} & \textbf{0.369} \\
\bottomrule
\end{tabular}}
\vspace{2pt}

\begin{minipage}{0.98\linewidth}
\footnotesize
\textit{Note.} For MLP-style adapters, ``N/A'' denotes no bottleneck, and all soft-token ablations use this no-bottleneck variant. For Q-Former adapters, $L$ denotes the number of Q-Former layers, $C$ denotes the context width, and $r / \alpha$ denotes the decoder-side LoRA rank and scaling factor. ``Ada.'' denotes the number of trainable adapter parameters.
\end{minipage}
\end{table}
\begin{table*}[t]
\centering
\setlength{\tabcolsep}{2.3pt}
\caption{
Main comparison across classification, fact retrieval, and gist summarization tasks.
We report ROUGE-L and BERTScore in the main paper as high-level generation quality metrics.
}
\label{tab:main_results}
\resizebox{\textwidth}{!}{
\begin{tabular}{llc|cc|cc|cc|cc}
\toprule
\multirow{2}{*}{\textbf{Method}} 
& \multirow{2}{*}{\textbf{Decoder}} 
& \multirow{2}{*}{\shortstack{\textbf{Injected}\\\textbf{Layer}}}
& \multicolumn{2}{c|}{\textbf{Classification}}
& \multicolumn{2}{c|}{\textbf{Fact}}
& \multicolumn{2}{c|}{\textbf{Gist}}
& \multicolumn{2}{c}{\textbf{Overall}} \\
\cmidrule(lr){4-5}
\cmidrule(lr){6-7}
\cmidrule(lr){8-9}
\cmidrule(lr){10-11}
& & 
& \textbf{R-L} & \textbf{BERTS}
& \textbf{R-L} & \textbf{BERTS}
& \textbf{R-L} & \textbf{BERTS}
& \textbf{R-L} & \textbf{BERTS} \\
\midrule

\multicolumn{11}{l}{\textit{Donor: Qwen3-4B-Instruct-2507}} \\
UAV 
& Qwen3-4B (Full) 
& 0
& \mstd{0.272}{0.146} & \mstd{0.315}{0.164}
& \mstd{0.233}{0.327} & \textbf{\mstd{0.374}{0.345}}
& \mstd{0.273}{0.177} & \mstd{0.323}{0.200}
& \mstd{0.254}{0.252} & \mstd{0.344}{0.271} \\

\rowcolor{lightblue} UAV 
& Qwen3-4B (AOT from Llama) 
& 0
& \textbf{\mstd{0.276}{0.143}} & \textbf{\mstd{0.323}{0.157}}
& \textbf{\mstd{0.239}{0.335}} & \mstd{0.372}{0.350}
& \textbf{\mstd{0.281}{0.191}} & \textbf{\mstd{0.326}{0.212}}
& \textbf{\mstd{0.260}{0.258}} & \textbf{\mstd{0.347}{0.274}} \\

AO 
& Qwen3-4B 
& 1
& \mstd{0.270}{0.147} & \mstd{0.319}{0.154}
& \mstd{0.129}{0.195} & \mstd{0.158}{0.287}
& \mstd{0.239}{0.152} & \mstd{0.270}{0.186}
& \mstd{0.198}{0.181} & \mstd{0.234}{0.242} \\

LatentQA 
& Qwen3-4B 
& 1
& \mstd{0.264}{0.135} & \mstd{0.308}{0.171}
& \mstd{0.200}{0.313} & \mstd{0.349}{0.338}
& \mstd{0.269}{0.175} & \mstd{0.309}{0.193}
& \mstd{0.235}{0.243} & \mstd{0.327}{0.266} \\

SelfIE 
& Qwen3-4B 
& 3
& \mstd{0.106}{0.047} & \mstd{-0.064}{0.092}
& \mstd{0.019}{0.034} & \mstd{-0.343}{0.135}
& \mstd{0.101}{0.052} & \mstd{-0.087}{0.117}
& \mstd{0.064}{0.060} & \mstd{-0.199}{0.180} \\

PatchScope 
& Qwen3-4B 
& 1
& \mstd{0.109}{0.049} & \mstd{-0.047}{0.090}
& \mstd{0.018}{0.035} & \mstd{-0.340}{0.133}
& \mstd{0.103}{0.053} & \mstd{-0.077}{0.115}
& \mstd{0.065}{0.062} & \mstd{-0.190}{0.183} \\

\midrule
\multicolumn{11}{l}{\textit{Donor: Llama-3.1-8B-Instruct}} \\

\rowcolor{lightblue} UAV 
& Llama-8B (Full) 
& 0
& {\mstd{0.286}{0.138}} & \textbf{\mstd{0.356}{0.155}}
& \textbf{\mstd{0.295}{0.369}} & \textbf{\mstd{0.421}{0.368}}
& \mstd{0.264}{0.166} & \mstd{0.317}{0.187}
& \textbf{\mstd{0.286}{0.275}} & \textbf{\mstd{0.379}{0.283}} \\

UAV 
& Qwen3-4B (Full) 
& 0
& {\mstd{0.289}{0.146}} & \mstd{0.353}{0.159}
& \mstd{0.265}{0.348} & \mstd{0.402}{0.356}
& \mstd{0.270}{0.179} & \textbf{\mstd{0.319}{0.207}}
& \mstd{0.274}{0.265} & \mstd{0.369}{0.278} \\

UAV 
& Qwen3-4B (AOT from Qwen) 
& 0
& \mstd{0.277}{0.135} & \mstd{0.321}{0.160}
& \mstd{0.240}{0.335} & \mstd{0.381}{0.351}
& {\mstd{0.266}{0.180}} & \mstd{0.308}{0.212}
& \mstd{0.257}{0.255} & \mstd{0.346}{0.276} \\

AO 
& Llama-8B 
& 1
& \mstd{0.288}{0.142} & \mstd{0.344}{0.155}
& \mstd{0.263}{0.352} & \mstd{0.397}{0.353}
& \mstd{0.271}{0.182} & \mstd{0.312}{0.203}
& \mstd{0.273}{0.267} & \mstd{0.362}{0.275} \\

LatentQA 
& Llama-8B 
& 1
& \textbf{\mstd{0.291}{0.146}} & \mstd{0.348}{0.160}
& \mstd{0.280}{0.367} & \mstd{0.408}{0.365}
& \textbf{\mstd{0.284}{0.178}} & \mstd{0.317}{0.193}
& \mstd{0.285}{0.276} & \mstd{0.370}{0.282} \\

SelfIE 
& Llama-8B 
& 3
& \mstd{0.101}{0.063} & \mstd{-0.006}{0.100}
& \mstd{0.022}{0.050} & \mstd{-0.285}{0.166}
& \mstd{0.113}{0.102} & \mstd{-0.005}{0.169}
& \mstd{0.066}{0.080} & \mstd{-0.136}{0.204} \\

PatchScope 
& Llama-8B 
& 1
& \mstd{0.120}{0.056} & \mstd{0.016}{0.083}
& \mstd{0.017}{0.035} & \mstd{-0.322}{0.147}
& \mstd{0.114}{0.086} & \mstd{-0.009}{0.135}
& \mstd{0.071}{0.075} & \mstd{-0.147}{0.207} \\

\bottomrule
\end{tabular}}
\vspace{2pt}

\begin{minipage}{0.98\textwidth}
\footnotesize
\textit{Note.}
All entries report the mean and standard deviation across individual evaluation examples. The standard deviation reflects per-example variation in open-ended generation rather than variation across random seeds. \textit{Full} denotes full two-stage adaptation on the current donor–decoder pair: Stage 1 trains the adapter with the decoder frozen, and Stage 2 jointly trains the adapter and decoder-side LoRA while keeping the decoder backbone frozen. AOT denotes adapter-only transfer. In both AOT settings, the current decoder is Qwen3-4B, and ``AOT from X'' indicates that the frozen decoder-side LoRA was learned through \textit{Full} adaptation using donor X. For the Qwen3-4B donor with AOT from Llama, we reuse the LoRA learned with the Llama-8B donor and Qwen3-4B decoder. For the Llama-8B donor with AOT from Qwen, we reuse the LoRA learned with the Qwen3-4B donor and Qwen3-4B decoder. In each case, only a new adapter for the current donor is trained.
\end{minipage}
\end{table*}

\subsection{Adapter Architectures for Activation-to-Text Alignment}\label{sec:arch}
To understand how adapter architecture affects UAV performance, we conduct separate design studies for MLP-style and Q-Former-style adapters. The result is shown in Table~\ref{tab:adapter_arch}. Full results are detailed in Appendix~\ref{app:architecture}.

\paragraph{MLP-style adapters.}
For MLP-style adapters, we vary the bottleneck size and the number of projected soft tokens while keeping the LoRA setup fixed.
Increasing the bottleneck dimension from 1024 to 3072 does not improve performance, and removing the bottleneck achieves the best validation loss among MLP variants.
This suggests that bottleneck compression may discard information useful for activation inversion.

We also find that increasing the number of projected soft tokens does not yield monotonic gains.
The 128-token variant produces twice as many soft tokens as the maximum input length, but has the largest adapter size and the worst validation loss.
Together, these results indicate that adding bottlenecks or excessive soft tokens does not naively improve verbalization, and too many soft tokens may introduce redundancy or noise and make optimization harder.
Therefore, we favor setting the soft-token length close to the maximum input length.

\paragraph{Q-Former adapters.}
For Q-Former adapters, we vary the number of cross-attention layers, context slots, and decoder-side LoRA capacity. 
Increasing depth from QF-Base to QF-Deep only yields marginal gains in generation metrics and does not reduce validation loss, showing that deeper adapter-side transformation alone is insufficient. 
Increasing context width from QF-Deep to QF-Wide gives more consistent improvements, indicating that a wider latent memory helps preserve activation information. 
The strongest result is achieved by QF-LoRA, which increases decoder-side LoRA capacity while using fewer adapter parameters than QF-Deep and QF-Wide. 
This suggests that, after the adapter aligns donor activations with the decoder embedding space, decoder-side adaptation is more important for converting the inverted activation signal into natural-language outputs.

\begin{table*}[t]
\centering
\setlength{\tabcolsep}{2pt}
\begin{minipage}{0.47\linewidth}
\caption{
Cross-donor verbalization with Qwen3-4B-Instruct-2507 as the shared decoder across all donor models.
We report validation loss, ROUGE-L, and BERTScore in the main paper.
}
\label{tab:cross_donor_main}
\scalebox{0.8}{
\begin{tabular}{lccc}
\toprule
\textbf{Donor}
& \textbf{Val $\downarrow$}
& \textbf{R-L $\uparrow$}
& \textbf{BERTS $\uparrow$} \\
\midrule
Llama-3.1-8B-Instruct
& 1.6388
& $0.274_{\pm 0.265}$
& $0.369_{\pm 0.278}$ \\

Gemma-3-4B-IT
& 1.6790
& $0.266_{\pm 0.263}$
& $0.354_{\pm 0.274}$ \\

Gemma-3-12B-IT
& 1.6589
& $0.269_{\pm 0.258}$
& $0.355_{\pm 0.281}$ \\

\rowcolor{blue!12}
Yi-1.5-34B-Chat
& \textbf{1.5837}
& $\mathbf{0.299}_{\pm 0.277}$
& $\mathbf{0.377}_{\pm 0.296}$ \\
\bottomrule
\end{tabular}
}\\[2pt]
\begin{minipage}{0.98\linewidth}
\footnotesize
\textit{Note.}
All rows use Qwen3-4B-Instruct-2507 as the shared decoder. Val denotes validation loss, R-L denotes ROUGE-L, and BERTS denotes BERTScore.
The best result per column is highlighted in bold.
\end{minipage}
\end{minipage}\hfill
\begin{minipage}{0.48\linewidth}
\caption{
Effect of decoder size on cross-decoding performance.
All results use Llama-3.1-8B-Instruct layer-27 activations as the donor and Qwen-family decoders of different sizes.
}
\label{tab:decoder_size_sweep}
\scalebox{0.82}{
\begin{tabular}{lccccc}
\toprule
\textbf{Decoder}
& \textbf{Val $\downarrow$}
& \textbf{Tok-F1 $\uparrow$}
& \textbf{R-L $\uparrow$}
& \textbf{chrF $\uparrow$}
& \textbf{BERTS $\uparrow$} \\
\midrule
Qwen3-0.6B
& 1.9172
& 0.250
& 0.239
& 0.248
& 0.323 \\

Qwen3-4B
& 1.6388
& 0.284
& 0.274
& 0.281
& 0.369 \\

Qwen3-8B
& 1.5634
& 0.294
& 0.283
& 0.291
& 0.368 \\

\rowcolor{blue!12}
Qwen3-14B
& \textbf{1.5050}
& \textbf{0.309}
& \textbf{0.296}
& \textbf{0.300}
& \textbf{0.388} \\
\bottomrule
\end{tabular}
}
\\[2pt]
\begin{minipage}{0.98\linewidth}
\footnotesize
\textit{Note.}
All rows use Llama-3.1-8B-Instruct layer-27 activations as the donor.
Val denotes the best fine-tuning validation loss; R-L denotes ROUGE-L; BERTS denotes BERTScore.
The best result in each column is highlighted in bold.
\end{minipage}
\end{minipage}
\end{table*}
We adopt QF-LoRA as the default setting for subsequent experiments, as it achieves the lowest validation loss and the strongest semantic similarity while maintaining better adapter-side parameter efficiency than the strongest MLP variant.

\subsection{Comparison with Activation Verbalization Baselines}\label{sec:baselines}

We evaluate UAV against representative activation verbalization baselines under a shared generative QA protocol covering classification, fact retrieval, and gist summarization.
These tasks require recovering information at different levels from a single donor activation, from label-level decisions to fine-grained facts and high-level summaries; examples are provided in Appendix~\ref{app:questions}.
We compare against training-free patching methods, including SelfIE and PatchScopes, as well as training-based self-explanation methods, including AO and LatentQA.
For training-based baselines, we use their native self-explanation setting, which is favorable because no cross-model alignment is required.
Thus, this comparison tests whether UAV can remain competitive with specialized self-explanation methods while supporting cross-model verbalization.

As shown in Table~\ref{tab:main_results}, under the self-explanation setting, UAV achieves strong performance against prior training-based approaches. 
For Qwen3-4B, our full two-stage adaptation obtains the best overall ROUGE-L and BERTScore among self-explaining methods, outperforming both AO and LatentQA. 
The gain is especially clear on fact retrieval, where UAV achieves substantially higher BERTScore than AO, suggesting that the learned adapter-decoder alignment better preserves fine-grained factual information encoded in the activation.
For Llama-3.1-8B, our full Llama-8B verbalizer achieves the strongest overall BERTScore and remains competitive with LatentQA on ROUGE-L, while outperforming AO on fact retrieval and overall semantic similarity. 
These results show that UAV is not merely a cross-model extension, but also a strong activation verbalizer in the conventional self-explanation setting. Qualitative examples in Table~\ref{tab:qualitative-examples} further illustrate successful factual recovery while highlighting metric brittleness and typical failure cases for short answers.

UAV shows a larger advantage over training-free patching methods.
SelfIE and PatchScopes obtain substantially lower scores across both donors, especially on fact retrieval, where their BERTScore values are near zero or negative.
This indicates that directly patching activations into prompts can recover limited topical cues but struggles with fine-grained factual and semantic information.
By learning to map donor activations into decoder-readable soft tokens, UAV enables the decoder to use the activation signal more reliably.
\begin{figure*}
    \includegraphics[width=0.33\linewidth]{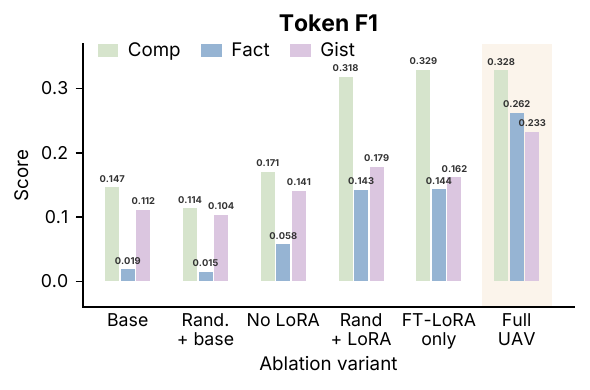}
    \includegraphics[width=0.33\linewidth]{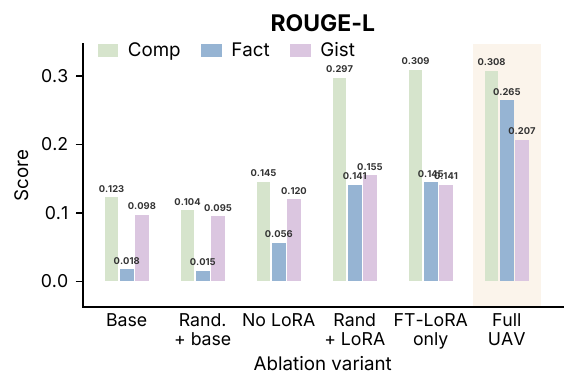}
    \includegraphics[width=0.33\linewidth]{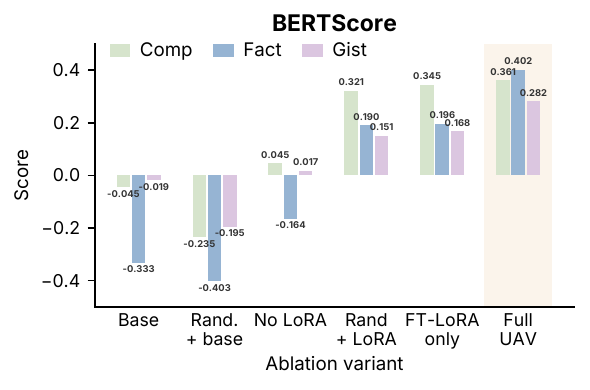}
    \caption{Disentangling the roles of the activation adapter and decoder-side LoRA.}\label{fig:activation}
\end{figure*}

Finally, UAV demonstrates strong transferability through adapter-only transfer. 
In this setting, the decoder-side LoRA learned from one donor is frozen, and only a new adapter is trained for the current donor. 
Despite this restricted training setup, AOT remains close to full two-stage adaptation. 
For Qwen3-4B, AOT even slightly outperforms the corresponding full self-explanation setting on the overall metrics. 
For Llama-3.1-8B, AOT shows only a moderate drop compared with full adaptation while still clearly outperforming training-free baselines and remaining competitive with training-based self-explanation methods. 
These results highlight a key advantage of UAV: it can reuse decoder-side explanation behavior across donors and adapt to a new donor by training only the adapter. A dataset-level analysis shows that UAV performs better on concrete
entity-centered retrieval than on abstract, sentiment, and semantically
flexible tasks; full per-dataset results and representative failure
cases are provided in Appendix~\ref{app:dataset_error_analysis}.

\subsection{Cross-Model Verbalization with a Shared Decoder}\label{sec:universal}
We study UAV's shared-decoder generalization across donor models by varying donor models and decoder sizes, and compare cross-model verbalization with self-decoding.
\paragraph{Shared Decoder Across Donor Models.} 
We further test whether a shared decoder can verbalize activations from different donor models.
As shown in Table~\ref{tab:cross_donor_main}, we fix Qwen3-4B-Instruct-2507 as the decoder and vary the donor across Llama, Gemma, and Yi families.
UAV achieves consistent verbalization performance across all tested heterogeneous donors, showing that the adapter-decoder interface is not tied to a single model family.

The results also suggest that decoding difficulty is not determined by model size alone.
While Gemma-3-12B improves over Gemma-3-4B, Llama-3.1-8B outperforms both Gemma donors, and Yi-1.5-34B achieves the best overall performance.
These comparisons indicate that cross-donor verbalization depends not only on donor scale, but also on donor-specific factors that affect alignment to the shared decoder space.
\begin{figure}[t]
    \centering
    \includegraphics[width=1\linewidth]{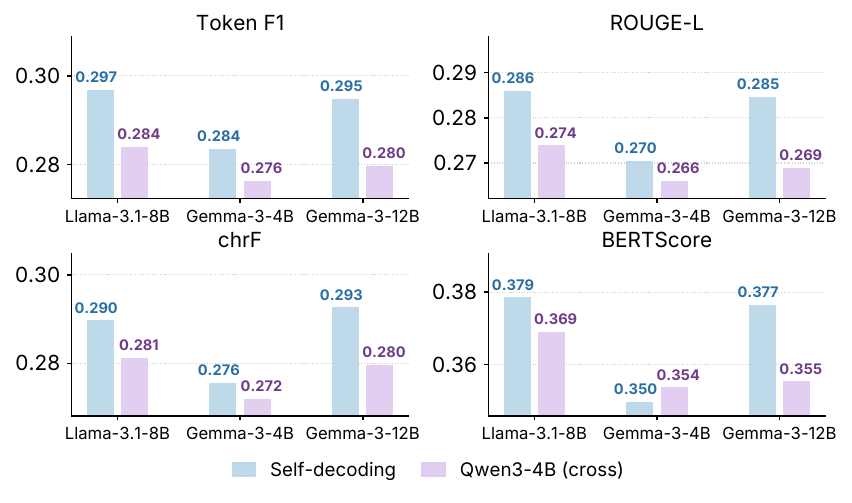}
    \caption{Overall performance comparison between self-decoding and cross-model verbalization.}
    \label{fig:cross}
\end{figure}

\paragraph{Effect of Decoder Size.} We study how decoder capacity affects cross-decoding by fixing the donor as Llama-3.1-8B-Instruct layer-27 activations and varying the Qwen-family decoder size. 
As shown in Table~\ref{tab:decoder_size_sweep}, increasing the decoder from 0.6B to 4B substantially improves both validation loss and generation quality. 
The 14B decoder achieves the best performance across all metrics, suggesting that larger decoders can better convert aligned activation signals into activation-conditioned natural-language outputs. 
However, the trend is not strictly monotonic: Qwen3-8B obtains a slightly lower validation loss than Qwen3-4B but performs worse on all generation metrics. 
This indicates that decoder scaling alone does not guarantee better activation verbalization, and that optimization dynamics or decoder-specific instruction-following behavior can also affect the final generation quality.

\paragraph{External Decoder vs. Self-Decoding}
We compare cross-model verbalization with self-decoding, where each donor also serves as its own decoder. As shown in Figure~\ref{fig:cross}, self-decoding generally achieves stronger results across most donors and metrics, suggesting that a model's own decoder is naturally better aligned with its internal activation space. In contrast, using Qwen3-4B as a shared external decoder requires the adapter to bridge heterogeneous representation spaces, making cross-model verbalization more challenging.
\begin{figure*}
\begin{minipage}{0.49\linewidth}
\subfloat[Layerwise]{\includegraphics[width=0.48\linewidth]{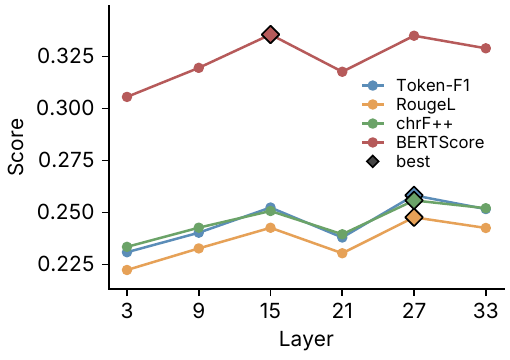}\label{fig:layers}}
\subfloat[Data Scale]{\includegraphics[width=0.48\linewidth]{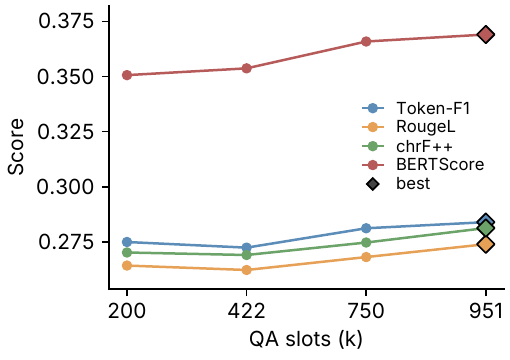}\label{fig:datascale}}
\caption{Additional ablation results on activation layer and Stage 2 training data scale.
(a) Layer-wise verbalizer performance using activations from different layers of Qwen3-4B.
(b) Effect of Stage 2 QA data scale on verbalization performance.}
\end{minipage}\hfill
\begin{minipage}{0.49\linewidth}
\subfloat[500K]{\includegraphics[width=0.48\linewidth]{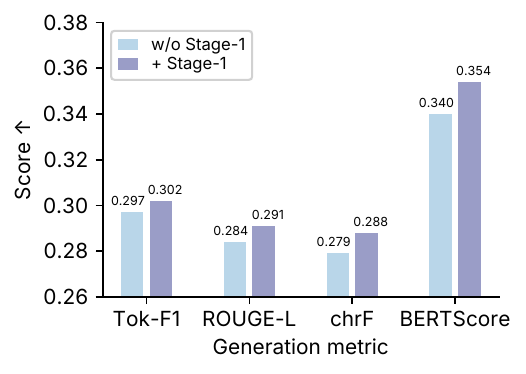}}
\subfloat[951K]{\includegraphics[width=0.48\linewidth]{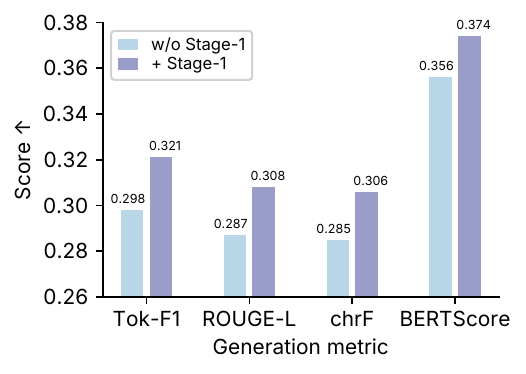}}
\caption{
Effect of Stage-1 pretraining on generation metrics.
We compare w/o Stage-1 and +Stage-1 under (a) 500k and (b) 951k Stage-2 training data.
Stage-1 pretraining consistently improves all generation metrics, especially under the full-data setting.
}
\label{fig:stage1_ablation}
\end{minipage}
\end{figure*}

Despite this disadvantage, the performance gap remains moderate. 
For Llama-3.1-8B and Gemma-3-12B, cross-model verbalization retains competitive overall scores across Token-F1, ROUGE-L, chrF, and BERTScore, showing that the shared Qwen3-4B decoder can still recover substantial information from non-Qwen activations. 
The gap is smallest for Gemma-3-4B, but the paired-bootstrap
confidence intervals include zero and the metric directions differ.
Overall, self- and cross-decoding are largely comparable. Task-level comparisons are provided in Appendix~\ref{app:external}

\subsection{Validating the Role of Activation Signal and Decoder Models}\label{sec:control}
We further disentangle the roles of the activation adapter and the decoder-side fine-tuning in UAV.
To this end, we compare controlled variants: 
(1) \textsc{Base}, which uses the base decoder without an adapter or decoder-side LoRA; 
(2) \textsc{Rand.+Base}, which injects randomly initialized adapter outputs into the base decoder; 
(3) \textsc{Stage-1}, which uses the adapter trained only by activation-to-text alignment with the base decoder; 
(4) \textsc{Rand. Align}, which trains with randomly mismatched activation-target pairs; 
(5) \textsc{FT-LoRA only}, which keeps the Stage-2 fine-tuned decoder-side LoRA but removes the activation-conditioned adapter outputs; and 
(6) \textsc{Full UAV}, which uses both the trained adapter and the fine-tuned decoder-side LoRA with correctly aligned activations. \textsc{Base} tests whether the instruction-tuned decoder can answer from its prior without donor activations.

Figure~\ref{fig:activation} shows that the base decoder alone performs poorly, especially on fact retrieval.
This indicates that the evaluation questions cannot be answered reliably from the decoder prior alone.
Adding a randomly initialized adapter to the base decoder does not improve the results, confirming that arbitrary soft-token injection cannot substitute for meaningful activation alignment.
The Stage-1 adapter provides a modest improvement, showing that activation-to-text alignment helps the decoder access some information from donor activations, although Stage-1 alignment alone remains insufficient for downstream explanation-oriented QA.

The comparison between \textsc{FT-LoRA only} and \textsc{Full UAV} further reveals complementary roles of decoder-side adaptation and the activation adapter.
The fine-tuned LoRA alone substantially improves comprehension-style performance, suggesting that Stage-2 decoder tuning helps the model learn the task format and classification-oriented decision patterns from the QA supervision.
However, \textsc{FT-LoRA only} remains much weaker on fact retrieval and gist summarization, where the model must recover input-specific information from the donor activation.
In contrast, \textsc{Full UAV} achieves the strongest performance on these information-intensive tasks, demonstrating that the trained adapter provides essential activation-conditioned evidence beyond what is captured by decoder-side LoRA or question priors.
These results suggest that decoder-side fine-tuning mainly improves instruction following and task-level behavior, while the adapter is critical for extracting fine-grained factual and semantic content from activations.

\subsection{Extending UAV to New Tasks}\label{sec:extend}
We study whether an existing UAV verbalizer can be extended from
classification and gist summarization to fact retrieval under the
500K-QA setting. In all settings, the Stage-1 adapter alignment is
reused without retraining. We compare sequential fine-tuning on the
new task with joint Stage-2 training on both the original and new tasks.

\begin{table}[t]
    \centering
    \small
    \caption{
       Extending UAV to fact retrieval under the 500K-QA setting. Each entry reports validation loss and BERTScore for each evaluation subset. Sequential fine-tuning is denoted by $\rightarrow$.
    }
    \label{tab:task_extension}
    \resizebox{\columnwidth}{!}{
    \begin{tabular}{lccc}
        \toprule
        \textbf{Setting}
        & \textbf{Class.+Gist}
        & \textbf{Fact}
        & \textbf{Overall} \\
        & \textbf{Loss / BERTS}
        & \textbf{Loss / BERTS}
        & \textbf{Loss / BERTS} \\
        \midrule

        Class.+Gist
        & 2.0294 / 0.314
        & 2.4856 / 0.143
        & 2.2402 / 0.234 \\

        Class.+Gist $\rightarrow$ Fact
        & 2.2542 / 0.195
        & \textbf{1.6343 / 0.366}
        & 1.9623 / 0.275 \\

        Joint training
        & \textbf{2.0249 / 0.314}
        & 1.6413 / 0.361
        & \textbf{1.8430 / 0.336} \\

        \bottomrule
    \end{tabular}
    }
\end{table}

Sequential fine-tuning substantially improves fact retrieval but
degrades performance on the original classification and gist tasks.
In contrast, joint training largely preserves the original-task
performance while nearly matching the fact-retrieval performance of
sequential fine-tuning. These results show that the Stage-1 alignment
can be reused when extending UAV to new tasks, while joint Stage-2
training helps reduce forgetting.

\subsection{Additional Ablations}\label{sec:ablations}

\paragraph{Activation Layer.}
We evaluate how the choice of activation layer affects verbalization performance by decoding activations from different depths of the donor model. 
Specifically, under the self-explanation setting, we study Layers 3, 9, 15, 21, 27, and 33 of Qwen3-4B. 
As shown in Figure~\ref{fig:layers}, performance generally improves from early to middle layers, with the strongest overall results around the middle-to-late layers. 
This suggests that these layers provide more informative and interpretable representations for activation verbalization, consistent with prior observations that later layers tend to encode richer and more abstract semantic knowledge~\cite{jin2024exploring}. 
Additional layer-wise ablation results are provided in Appendix~\ref{app:layerwise}.

\paragraph{Training Data Scale.}
We study the effect of Stage 2 data scale by training with 200K, 422K, 750K, and 951K query-oriented QA pairs.
Figure~\ref{fig:datascale} shows that larger and more diverse QA supervision improves downstream activation interpretation, so we use 951K QA pairs as the default scale.
Additional training curves and per-metric results are provided in Figure~\ref{fig:app-datascale}.

\paragraph{Activation-to-Text Warmup.}
We further study the role of the Stage-1 activation-to-text alignment stage. 
Stage-1 serves as a warmup that first aligns donor activations with the decoder embedding space before query-oriented fine-tuning. 
As shown in Figure~\ref{fig:stage1_ablation}, removing this warmup consistently hurts performance across both training scales, while also increasing validation loss and perplexity. 
The improvement is more pronounced under the full training setting, suggesting that Stage-1 provides a better initialization for subsequent activation-grounded QA. 
These results show that activation-to-text alignment is an important prerequisite for reliable verbalizer training.

\section{Conclusions and Future Work}\label{sec:conclusion}

We present \textsc{Universal Activation Verbalizer} (UAV), an adapter-decoder framework that explains activations from heterogeneous donor models with a shared decoder. 
UAV remains competitive with self-explanation baselines while enabling cross-model verbalization across model families and scales. 
Ablations show that decoder-side tuning improves task behavior, whereas the adapter provides activation-grounded factual and semantic information. 
Future work will improve fact retrieval and gist understanding through stronger activation alignment, multi-layer features, and richer factual or summarization supervision.

\section*{Limitations}
Although UAV enables cross-model activation verbalization, several limitations remain.
First, UAV remains limited in fact retrieval and gist summarization, especially when recovering fine-grained attributes or abstract semantic content from a single activation.
Second, our current setting focuses on short-context inputs, leaving long-context and document-level activation explanation for future work.
Third, adapter-only transfer still requires training a donor- and layer-specific adapter, which may be costly for large-scale layer-wise analysis.
Lastly, the source-grounded evaluation relies on LLM judges rather than human annotators.
\bibliography{ref}

\clearpage
\appendix

\section{Additional Method Details}\label{app:methods}
\subsection{Teaching Forcing}\label{app:teacher-force}

We use teacher forcing for both the activation-to-text alignment stage and the explanation-oriented instruction-tuning stage.
Given an input token sequence $\mathbf{t}_{0:k}=\langle t_0,\ldots,t_k\rangle$ and a donor activation $\mathbf{h}^{\ell}_{i}$ from layer $\ell$ at position $i$, the adapter first maps the activation into $n$ decoder-readable soft tokens
$\mathbf{z}_{1:n}=A_{\phi}(\mathbf{h}^{\ell}_{i})$.
During Stage~1, the target output is the prefix $\mathbf{t}_{0:i}$.
At decoding step $j$, instead of conditioning on previously generated tokens $\hat{\mathbf{t}}_{<j}$, we feed the ground-truth prefix $\mathbf{t}_{<j}$ to the verbalizer.
The conditional likelihood is therefore
\begin{equation}
p_{\theta,\phi}(\mathbf{t}_{0:i}\mid \mathbf{h}^{\ell}_{i})
=
\prod_{j=0}^{i}
p_{\theta,\phi}
\left(
t_j
\mid
\mathbf{z}_{1:n}, \mathbf{t}_{<j}
\right),
\end{equation}
where $\phi$ denotes adapter parameters and $\theta$ denotes the trainable decoder-side parameters.
In Stage~1, $\theta$ is frozen and only $\phi$ is optimized.

The Stage~1 teacher-forcing objective is the token-level negative log-likelihood:
\begin{equation}
\mathcal{L}_{\mathrm{rec}}
=
-\frac{1}{i+1}
\sum_{j=0}^{i}
\log
p_{\theta,\phi}
\left(
t_j
\mid
\mathbf{z}_{1:n}, \mathbf{t}_{<j}
\right).
\end{equation}

For Stage~2, each training example contains a textual prompt or question $\mathbf{x}$ and a target answer sequence
$\mathbf{y}_{1:T}=\langle y_1,\ldots,y_T\rangle$.
The soft tokens are prepended to the prompt embeddings, and the verbalizer predicts the answer autoregressively.
Under teacher forcing, the model conditions on the ground-truth answer prefix $\mathbf{y}_{<j}$ at each decoding step:
\begin{equation}
p_{\theta,\phi}(\mathbf{y}_{1:T}\mid \mathbf{h}^{\ell}_{i}, \mathbf{x})
=
\prod_{j=1}^{T}
p_{\theta,\phi}
\left(
y_j
\mid
\mathbf{z}_{1:n}, \mathbf{x}, \mathbf{y}_{<j}
\right).
\end{equation}
The explanation-oriented loss is
\begin{equation}
\mathcal{L}_{\mathrm{exp}}
=
-\frac{1}{T}
\sum_{j=1}^{T}
\log
p_{\theta,\phi}
\left(
y_j
\mid
\mathbf{z}_{1:n}, \mathbf{x}, \mathbf{y}_{<j}
\right).
\end{equation}

Thus, teacher forcing replaces the model-generated history with the corresponding ground-truth history during training.
This stabilizes optimization by exposing the adapter and decoder to correct prefixes at every step, while inference remains fully autoregressive.

\subsection{Adapter Designs}\label{app:adapter-design}
We adopted two different adapter designs to study which is more efficient under the setting of UAV.
\paragraph{MLP-base design.} 
Given an input activation $\boldsymbol{h} \in \mathbb{R}^{d_{\mathrm{Donor}}}$, 
the MLP adapter first applies layer normalization and projects the activation 
into a hidden space. By default, we set the hidden dimension of $\boldsymbol{z}_{1}$ to $d_1 = 2d_\mathrm{Donor}$, 
unless an explicit bottleneck dimension $b$ is used.
\begin{equation}
\label{eq:mlp}
\begin{aligned}
\tilde{\boldsymbol{h}} 
&= \mathrm{LN}(\boldsymbol{h}), \\
\boldsymbol{z}_1 
&= \mathrm{Dropout}\!\left(
    \mathrm{GELU}(W_1 \tilde{\boldsymbol{h}} + \boldsymbol{b}_1)
\right), \\
\boldsymbol{z}_b 
&= \mathrm{Dropout}\!\left(
    \mathrm{GELU}(W_b \boldsymbol{z}_1 + \boldsymbol{b}_b)
\right), \\
\boldsymbol{y} 
&= W_{\mathrm{out}} \boldsymbol{z}_{\star} + \boldsymbol{b}_{\mathrm{out}}, \\
Y 
&= \mathrm{reshape}(\boldsymbol{y};\, n \times d).
\end{aligned}
\end{equation}

Here, the bottleneck transformation that produces $\boldsymbol{z}_b$ is optional. 
When the bottleneck layer is disabled, we set 
$\boldsymbol{z}_{\star}=\boldsymbol{z}_1$ and $d_{\star}=d_1$; 
otherwise, we set $\boldsymbol{z}_{\star}=\boldsymbol{z}_b$ and $d_{\star}=b$.
The projection matrices are 
$W_1 \in \mathbb{R}^{d_h \times d_{\mathrm{Donor}}}$, 
$W_b \in \mathbb{R}^{b \times d_h}$, and 
$W_{\mathrm{out}} \in \mathbb{R}^{(n \cdot d) \times d_{\star}}$.

\paragraph{Attention-layer-based design.} Given an input activation $\boldsymbol{h} \in \mathbb{R}^{d_{\mathrm{Donor}}}$, 
the cross-attention adapter follows a Q-Former-style design. 
It first expands the donor activation into a set of latent context slots, 
and then uses learnable query tokens to extract information from these slots through cross-attention. 
The final query states are used as output soft tokens in the decoder embedding space.

We first project the donor activation into $M$ context slots in the decoder embedding space. 
These slots serve as a latent memory that stores different views of the donor activation.
\begin{equation}
\label{eq:cross-context-slots}\small
C 
= 
\mathrm{LN}_{c}\!\left(
    \mathrm{reshape}
    \left(
        W_{\mathrm{ctx}}\boldsymbol{h} + \boldsymbol{b}_{\mathrm{ctx}};\,
        M \times d
    \right)
\right),
\end{equation}
where $C \in \mathbb{R}^{M \times d}$ denotes the constructed context slots, 
$M$ is the number of context slots, and $d$ is the decoder embedding dimension. 
The projection matrix is 
$W_{\mathrm{ctx}} \in \mathbb{R}^{(M d) \times d_{\mathrm{Donor}}}$, 
with bias $\boldsymbol{b}_{\mathrm{ctx}} \in \mathbb{R}^{Md}$. 
Intuitively, this step maps a single donor activation vector into a sequence of $M$ latent tokens, 
which can later be attended to by the learnable queries.

After constructing the context slots, we introduce $n$ learnable query tokens. 
Each query token corresponds to one output soft token.
\begin{equation}
\label{eq:cross-query-init}
Q^{(0)} = Q_0,
\qquad
Q_0 \in \mathbb{R}^{n \times d}.
\end{equation}
Here, $n$ is the number of output soft tokens. 
Unlike the context slots $C$, which are input-dependent, 
$Q_0$ is a learnable parameter shared across inputs. 
Through cross-attention, these queries become conditioned on the current donor activation.

We stack $L$ cross-attention blocks. 
At layer $\ell$, the query states $Q^{(\ell-1)}$ attend to the context slots $C$. 
The cross-attention sublayer uses pre-normalization and a residual connection.
\begin{equation}
\label{eq:cross-attn-block}
\small\begin{aligned}
\hat{Q}^{(\ell)} 
&= \mathrm{LN}_{q}\!\left(Q^{(\ell-1)}\right), 
\qquad
\hat{C} = \mathrm{LN}_{c}(C), \\
Q'^{(\ell)} 
&= W_q \hat{Q}^{(\ell)}, 
\qquad
[K^{(\ell)}, V^{(\ell)}] = W_{kv}\hat{C}, \\
Q_h^{(\ell)}, K_h^{(\ell)}, V_h^{(\ell)}
&= 
\mathrm{split\text{-}heads}
\left(
    Q'^{(\ell)}, K^{(\ell)}, V^{(\ell)}
\right), \\
A^{(\ell)}
&=
\mathrm{Dropout}\!\left(
    \mathrm{softmax}
    \left(
        \frac{
        Q_h^{(\ell)} {K_h^{(\ell)}}^{\top}
        }{\sqrt{d_h}}
    \right)
\right), \\
O^{(\ell)}
&=
\mathrm{merge\text{-}heads}
\left(
    A^{(\ell)} V_h^{(\ell)}
\right), \\
\tilde{Q}^{(\ell)}
&=
Q^{(\ell-1)}
+
\mathrm{Dropout}
\left(
    W_o O^{(\ell)}
\right).
\end{aligned}
\end{equation}

We use $H$ attention heads and set $d_h=d/H$. 
After splitting into heads, 
$Q_h^{(\ell)} \in \mathbb{R}^{H \times n \times d_h}$ and 
$K_h^{(\ell)}, V_h^{(\ell)} \in \mathbb{R}^{H \times M \times d_h}$. 
Therefore, the attention map satisfies 
$A^{(\ell)} \in \mathbb{R}^{H \times n \times M}$, 
which indicates how each query token attends to the $M$ context slots. 
The projection matrices are 
$W_q \in \mathbb{R}^{d \times d}$, 
$W_{kv} \in \mathbb{R}^{2d \times d}$, and 
$W_o \in \mathbb{R}^{d \times d}$.

Each cross-attention block can optionally include a feed-forward sublayer. 
When $r=\texttt{ffn\_mult}>0$, we update the query states as
\begin{equation}
\label{eq:cross-ffn}\small
Q^{(\ell)}
=
\tilde{Q}^{(\ell)}
+
\mathrm{Dropout}\!\left(
    W_2
    \mathrm{GELU}
    \left(
        W_1
        \mathrm{LN}
        \left(
            \tilde{Q}^{(\ell)}
        \right)
    \right)
\right),
\end{equation}
where 
$W_1 \in \mathbb{R}^{rd \times d}$ and 
$W_2 \in \mathbb{R}^{d \times rd}$. We set $\texttt{ffn\_mult}=4$ for all our Q-Former-based adapters, as we found that the FFN sublayer substantially improves the performance of our verbalizer. 
When $\texttt{ffn\_mult}=0$, the FFN sublayer is disabled and we set
\begin{equation}
\label{eq:cross-no-ffn}
Q^{(\ell)} = \tilde{Q}^{(\ell)}.
\end{equation}

After $L$ cross-attention blocks, the final query states are normalized and used as the adapter output:
\begin{equation}
\label{eq:cross-output}\small
Y
=
\mathrm{Dropout}
\left(
    \mathrm{LN}_{\mathrm{out}}
    \left(
        Q^{(L)}
    \right)
\right),
\ 
Y \in \mathbb{R}^{n \times d}.
\end{equation}

\begin{table*}[t]
\centering
\setlength{\tabcolsep}{4pt}
\caption{Composition of the diversified training corpus. All 17 sources are used for
\textbf{Stage 1} activation-to-text reconstruction, while the 15 sources with
synthesized QA pairs are also used for \textbf{Stage 2} generative QA finetuning.
\textbf{Texts} denotes the number of source passages retained after sentence
packing, length filtering (10--64 Qwen3-4B tokens), and deduplication.
\textbf{QA items} denotes the number of flattened $(\text{activation}, Q, A)$
training triples derived from LLM-generated QA records; ``-'' indicates
sources used only in Stage 1.}
\label{tab:datasets}
\scalebox{0.81}{\begin{tabular}{llrlr}
\toprule
\textbf{Source} & \textbf{Domain} & \textbf{Texts} & \textbf{QA mode} & \textbf{QA items} \\
\midrule
\multicolumn{5}{l}{\textbf{Stage 1 (alignment / activation--text reconstruction): 17 sources}} \\
\addlinespace
\multicolumn{5}{l}{\textit{Wikipedia (first-paragraph snippets, factual QA)}} \\
\quad Person       & Biography           &  7{,}250 & factual       &  27{,}898 \\
\quad Place        & Geography           &  7{,}250 & factual       &  27{,}773 \\
\quad Event        & Events              &  7{,}250 & factual       &  27{,}808 \\
\quad Organization & Institutions        &  7{,}250 & factual       &  27{,}887 \\
\quad Work         & Books / films       &  7{,}250 & factual       &  27{,}892 \\
\quad Concept      & Abstract concepts   &  7{,}250 & factual       &  27{,}888 \\
\quad Generic      & Miscellaneous       &  7{,}250 & factual       &  27{,}830 \\
\addlinespace
\multicolumn{5}{l}{\textit{Long-form and classification (comprehension QA)}} \\
\quad peS2o (Scientific)            & Sci.\ abstracts & 50{,}250 & comprehension & 186{,}408 \\
\quad AG News                       & News headlines  & 50{,}250 & comprehension & 199{,}707 \\
\quad SST-2$^{\dagger}$             & Movie reviews   & 18{,}254 & comprehension &  62{,}876 \\
\quad TweetEval-Sentiment$^{\dagger}$ & Tweets        & 20{,}250 & comprehension &  78{,}225 \\
\quad TweetEval-Emotion$^{\dagger}$ & Tweets          &  3{,}074 & comprehension &  11{,}015 \\
\quad dair-ai/emotion               & Tweets          & 19{,}912 & comprehension &  77{,}521 \\
\quad LatentQA-Control$^{\dagger}$  & Diverse short text & 16{,}490 & comprehension & 64{,}662 \\
\quad MD-Gender$^{\dagger}$         & Gendered text   & 29{,}857 & -           & -       \\
\quad NER$^{\dagger}$               & Entity spans    & 14{,}923 & -           & -       \\
\addlinespace
\multicolumn{5}{l}{\textit{User queries (intent / query QA)}} \\
\quad LMSYS-Chat-1M (1st user turn) & Chatbot requests & 20{,}000 & query        &  75{,}326 \\
\addlinespace
\multicolumn{5}{l}{\textit{Stage-1-only deduplicated supplements (no synthetic QA)}} \\
\quad AG News (extra)               & News headlines  & 50{,}000 & -          & -       \\
\quad TweetEval-Sentiment (extra)   & Tweets          & 25{,}299 & -          & -       \\
\quad dair-ai/emotion (extra)       & Tweets          & 50{,}000 & -          & -       \\
\quad LMSYS-Chat-1M (extra)         & Chatbot requests & 50{,}000 & -          & -       \\
\midrule
\multicolumn{2}{l}{\textbf{Stage 1 alignment total (17 sources)}}
                                                      & \textbf{469{,}309} &      & -       \\
\midrule
\multicolumn{5}{l}{\textbf{Stage 2 (generative QA finetune): 15 sources $\subseteq$ Stage 1}} \\
\multicolumn{2}{l}{\textbf{Stage 2 QA finetune total}}
                                                      & 244{,}662          &      & \textbf{950{,}716} \\
\bottomrule
\end{tabular}}
\\[2pt]
{\parbox{\linewidth}{
\footnotesize $^{\dagger}$ Sources labeled as reused are drawn from the Activation Oracle
corpus~\cite{karvonen2025activation}. For each source, the final
$\min(250, 20\%)$ rows are held out for evaluation, and reported counts
include this evaluation tail. The four \textit{extra} blocks are disjoint
samples from the corresponding upstream corpora, with all Stage-2 texts
removed via hash- or index-based deduplication to prevent leakage between
Stage~1 activation--text alignment and Stage~2 QA finetuning. QA items are
generated offline using Qwen3-14B served by vLLM.
}}
\end{table*}

In this design, the context slots $C$ are input-dependent and provide a latent memory of the donor activation, 
whereas the query tokens $Q_0$ are input-independent learnable slots that define the output token positions. 
Through cross-attention, each query token learns to retrieve information from the context slots, 
resulting in activation-conditioned soft tokens. 
Compared with the MLP adapter, this design decouples the number of output tokens $n$ from the main projection parameters: 
increasing $n$ mainly enlarges the learnable query matrix $Q_0$, while the context projection and attention parameters remain unchanged.

\section{Datasets}\label{app:datasets}
The detailed combinations of datasets are shown in Table~\ref{tab:datasets}. 

\subsection{Data Sourcing}\label{app:curation}
\paragraph{Data Curation.}
To train a universal activation verbalizer, we construct a diversified short-context corpus that covers factual knowledge, semantic comprehension, classification-oriented attributes, and natural user intents.
Our goal is not to train a long-document reader, but to expose the verbalizer to diverse semantic phenomena under a controlled input length budget where most sequences are within 64 tokens.
The full data construction pipeline consists of raw text sourcing, length normalization, train/evaluation split construction, offline QA generation, and activation representation caching.

\paragraph{Raw Text Sources.}
We collect raw texts from 17 sources spanning three regimes.
First, we use Wikipedia entries to cover factual knowledge, grouping pages into seven subtypes: person, place, event, organization, work, concept, and generic.
Second, we include scientific abstracts, news, sentiment, emotion, gender, named-entity, and LatentQA-style control datasets to cover short-context semantic and classification-oriented attributes.
Third, we sample natural user requests from LMSYS-Chat-1M by retaining only the first user turn from each conversation after language, moderation, length, PII, and deduplication filtering.
All examples are normalized into a unified schema containing the text, source, subtype, and metadata fields.

\paragraph{Length Normalization.}
We normalize all raw texts into short activation contexts before caching representations or generating QA supervision.
For longer examples, we count tokens with the Qwen3-4B tokenizer, which is also the target model used for activation caching.
If an example exceeds 64 tokens, we first truncate it to the leading sentence.
The sentence splitter searches for real sentence boundaries while skipping common abbreviations such as ``Dr.'', ``U.S.'', and ``Ph.D.''.
If no reliable sentence boundary is found, we fall back to the leading paragraph prefix.
The truncated text is then re-validated, and examples that still violate the 30-character or 64-token constraint are discarded.

Some sources require a wider parsing window to extract metadata, but this window is never used as model input.
For example, Wikipedia examples are parsed with a longer lead window to identify entity descriptors and assign factual subtypes, while classification-style sources may use a wider window to infer labels or entity spans.
These metadata fields are stored only as prompt-side hints for QA generation.
The actual \texttt{text} field still comes from the globally normalized record and remains bounded by 64 Qwen3-4B tokens.
For LMSYS user queries, we additionally apply a source-specific character filter before global normalization, keeping only first-turn user requests between 30 and 400 characters and removing redaction artifacts that start with \texttt{NAME\_}.

This design makes the activation cache predictable and keeps QA generation lightweight.
Since all cached inputs are token-bounded under the same tokenizer as the target model, representation shards have a stable memory footprint.
Moreover, because every input text is at most 64 tokens, the full QA-generation prompt remains well below the vLLM context limit, even after adding source-specific instructions or label-focus hints.
Overall, QA generation, activation caching, and dataloader joining operate on the same normalized short-text distribution.

\begin{table}[t]
\centering
\setlength{\tabcolsep}{4pt}
\caption{
Source distribution of the test set.
Each source contributes 100 examples. Non-Wikipedia sources contain comprehension-style and gist questions,
while Wikipedia sources contain fact-retrieval questions.
}
\label{tab:app:testset}
\scalebox{0.65}{
\begin{tabular}{llcccc}
\toprule
\textbf{Category} & \textbf{Source} & \textbf{Gist} & \textbf{Comp.} & \textbf{Fact} & \textbf{Total} \\
\midrule
Topic / news & ag\_news & 25 & 75 & -- & 100 \\
\midrule
\multirow{2}{*}{Sentiment}
& sst2 & 28 & 72 & -- & 100 \\
& tweeteval\_sentiment & 25 & 75 & -- & 100 \\
\midrule
\multirow{2}{*}{Emotion}
& tweeteval\_emotion & 25 & 75 & -- & 100 \\
& dair\_emotion & 26 & 74 & -- & 100 \\
\midrule
User intent & lmsys\_user & 26 & 74 & -- & 100 \\
Behavior / control & latentqa\_control & 25 & 75 & -- & 100 \\
Scientific abstract & scientific & 27 & 73 & -- & 100 \\
\midrule
\multirow{7}{*}{Wikipedia}
& wikipedia\_concept & -- & -- & 100 & 100 \\
& wikipedia\_event & -- & -- & 100 & 100 \\
& wikipedia\_generic & -- & -- & 100 & 100 \\
& wikipedia\_organization & -- & -- & 100 & 100 \\
& wikipedia\_person & -- & -- & 100 & 100 \\
& wikipedia\_place & -- & -- & 100 & 100 \\
& wikipedia\_work & -- & -- & 100 & 100 \\
\midrule
\textbf{Total} & \textbf{15 sources} & \textbf{207} & \textbf{593} & \textbf{700} & \textbf{1500} \\
\bottomrule
\end{tabular}
}
\end{table}

\subsection{QA Pair Construction.}\label{sec:qa-pairs}
Given each passage, we generate question–answer pairs that supervise the adapter on different facets of the source content. We construct a corpus of approximately 838K passages spanning 17 subtypes drawn from heterogeneous sources: seven Wikipedia subtypes (person, place, event, concept, organization, work, generic), scientific abstracts, AG News, SST-2, TweetEval (sentiment, emotion, stance), DAIR-emotion, MD-Gender, an NER corpus, and a LatentQA control set. All passages are truncated to $\leq$64 tokens (Qwen tokenizer) and filtered to $\geq$30 characters so that every input fits a uniform context budget. Each passage is routed to one of two generation modes according to its subtype.

For Wikipedia passages we use a factual mode in which the generator produces up to four atomic QA pairs, each targeting a single attribute. A subtype-specific \texttt{fact\_hint} directs the generator toward attribute classes that are typical for the entity type — for example, nationality, profession, dates, affiliations, relationships for person; founding date, location, purpose, leadership for organization; creator, year, genre, cast, setting, reception for work; and dates, participants, locations, causes, outcomes for event. To prevent trivially solvable questions, the prompt additionally enforces three anti-leak constraints: questions must refer to the subject only by a generic placeholder ("this person", "this place", "this entity", etc.) rather than by its proper name; questions must not exceed 15 words; and questions must not quote four or more consecutive words from the passage.

For the remaining subtypes we use a comprehension mode that yields one one-sentence gist together with one to three QA pairs covering higher-level aspects such as the main event, topic, intent, tone, sentiment, emotion, or speaker attitude. Each comprehension subtype carries a custom extra\_hints field that names the dimensions of interest — for instance, aspects praised or criticized, overall verdict for SST-2, trigger or object of the feeling, overall mood for the emotion sources, and subject of debate, what the speaker argues for TweetEval-stance. The prompt explicitly bans generic template questions ("What is the overall tone?", "What is the intent of the text?") and requires answers to be expressed through concrete, content-grounded phrases lifted from the passage.

For labeled subtypes (SST-2, AG News, TweetEval-sentiment/emotion/stance, DAIR-emotion, MD-Gender, NER, LatentQA-control) the categorical label is supplied to the generator only as a focus hint identifying which dimension at least one QA pair must address — e.g., the speaker's attitude or evaluative tone for SST-2, the subject area or field of the news reported for AG News, the emotional state expressed by the speaker for DAIR-emotion, the position taken on the subject for stance, and the specific named entity mentioned in the text for NER. The prompt forbids the answer from copying the label string or close synonyms, requiring instead that the relevant property be described through distinctive phrases from the passage. This design reduces label leakage and prevents the adapter from learning a shortcut that maps activations directly to dataset labels.

Comprehension-mode examples are augmented with gist supervision. The gist answer is generated once as a standalone one-sentence paraphrase; at training time it is paired with a question sampled uniformly from a pool of 51 semantically equivalent summarization prompts (e.g., "What is the main idea of this text?", "Summarize this text in one sentence.", "State the gist of this text."). This question diversification prevents the adapter from overfitting to a fixed query template and encourages stable behavior under surface-level prompt variation.

All QA pairs are generated by Qwen3-14B-AWQ served through vLLM (batch size 32, max new tokens 300, JSON-constrained decoding). The full corpus is partitioned into uniform 20K-row shards across subtypes (~42 shards) and each shard's raw model outputs are persisted to disk before parsing, so that prompt or schema revisions can be applied offline without re-running generation.

\subsection{Concrete Examples for QA Tasks}\label{app:questions}
Table~\ref{tab:app:testset} summarizes the composition of our test set.
We sample 100 examples from each of the 15 sources, resulting in 1,500 evaluation examples in total.
For non-Wikipedia sources, we construct both comprehension-style questions and gist summarization questions.
For Wikipedia-based sources, we focus on fact-retrieval questions that ask for fine-grained attributes from the
input text. This design allows the evaluation set to cover three complementary levels of activation-grounded
understanding: task-oriented comprehension, high-level gist summarization, and fine-grained factual retrieval.

\section{Evaluation Metrics}\label{app:metrics}
We use five automatic metrics to evaluate the agreement between generated answers and references.

\paragraph{Token-level F1.}
Token-level F1 measures lexical overlap between the generated answer and the reference at the token level, following the standard QA evaluation protocol~\cite{rajpurkar-etal-2016-squad}. 
Given precision $P$ and recall $R$ over shared tokens, the score is computed as:
\begin{equation}
\mathrm{F1} = \frac{2PR}{P + R}.\label{eq:tokenf1}
\end{equation}

\begin{table}[t]
\centering
\setlength{\tabcolsep}{3pt}
\caption{
Adapter architecture exploration.
We organize the ablations into two adapter families: MLP-style adapters and Q-Former adapters.
For MLP adapters, we vary the bottleneck size and the number of projected soft tokens.
For Q-Former adapters, configurations are denoted by $(L, C, r/\alpha)$.
}
\label{tab:app-adapter-arch}
\scalebox{0.72}{
\begin{tabular}{llcccc}
\toprule
\textbf{ID} & \textbf{Configuration}
& \textbf{Ada.} & \textbf{LoRA}
& \textbf{Tok-F1 $\uparrow$} 
& \textbf{chrF++ $\uparrow$} \\
\midrule
\multicolumn{6}{l}{\textbf{MLP-style adapters}} \\
\multicolumn{6}{l}{\textit{Bottleneck size}} \\
MLP-B1 & 1024 & 195M & 132.1M & 0.270 & 0.264\\
MLP-B2 & 2048 & 367M & 132.1M & 0.269 & 0.267\\
MLP-B3 & 3072 & 540M & 132.1M & 0.272 & 0.268\\
\rowcolor{lightblue}
MLP-B4 & N/A  & 860M & 132.1M & 0.308 & 0.297\\
\multicolumn{6}{l}{\textit{Number of projected soft tokens}} \\
MLP-T1 & 8   & 99.7M  & 132.1M & 0.276 & 0.271 \\
MLP-T2 & 16  & 162.6M & 132.1M & 0.271 & 0.266\\
MLP-T3 & 64  & 540M   & 132.1M & 0.272 & 0.268\\
MLP-T4 & 128 & 1044M  & 132.1M & 0.260 & 0.259\\

\midrule
\multicolumn{6}{l}{\textbf{Q-Former adapters}} \\
QF-Base & $(1, 8, 16/32)$  & 162.8M & 33.0M  & 0.267 & 0.261\\
QF-Deep & $(4, 8, 16/32)$  & 398.8M & 33.0M & 0.272 & 0.268\\
QF-Wide & $(4, 16, 16/32)$ & 482.7M & 33.0M & 0.278 & 0.273\\
\rowcolor{lightblue}
QF-LoRA & $(2, 8, 64/128)$ & 241.4M & 132.1M &  0.321 & 0.306\\
\bottomrule
\end{tabular}}
\vspace{2pt}
\begin{minipage}{0.95\linewidth}
\footnotesize
\textit{Note.} For Q-Former adapters, $L$ denotes the number of Q-Former layers, 
$C$ denotes the context width, and $r/\alpha$ denotes the decoder-side LoRA rank and scaling factor.
``Ada.'' denotes the number of trainable adapter parameters.
\end{minipage}
\end{table}
\begin{table*}[t]
\centering
\setlength{\tabcolsep}{2.3pt}
\caption{
Additional comparison across classification, fact retrieval, and gist summarization tasks.
We report Token-F1 and chrF as complementary token-level and character-level generation metrics.
}
\label{tab:main_results_token_chrf}
\resizebox{\textwidth}{!}{
\begin{tabular}{llc|cc|cc|cc|cc}
\toprule
\multirow{2}{*}{\textbf{Method}} 
& \multirow{2}{*}{\textbf{Decoder}} 
& \multirow{2}{*}{\shortstack{\textbf{Injected}\\\textbf{Layer}}}
& \multicolumn{2}{c|}{\textbf{Classification}}
& \multicolumn{2}{c|}{\textbf{Fact}}
& \multicolumn{2}{c|}{\textbf{Gist}}
& \multicolumn{2}{c}{\textbf{Overall}} \\
\cmidrule(lr){4-5}
\cmidrule(lr){6-7}
\cmidrule(lr){8-9}
\cmidrule(lr){10-11}
& & 
& \textbf{Tok-F1} & \textbf{chrF}
& \textbf{Tok-F1} & \textbf{chrF}
& \textbf{Tok-F1} & \textbf{chrF}
& \textbf{Tok-F1} & \textbf{chrF} \\
\midrule

\multicolumn{11}{l}{\textit{Donor: Qwen3-4B-Instruct-2507}} \\

UAV 
& Qwen3-4B (Full) 
& 0
& \mstd{0.291}{0.148} & \mstd{0.271}{0.122}
& \mstd{0.229}{0.326} & \mstd{0.244}{0.289}
& \mstd{0.297}{0.181} & \mstd{0.283}{0.152}
& \mstd{0.263}{0.254} & \mstd{0.261}{0.221} \\

\rowcolor{lightblue} UAV 
& Qwen3-4B (AOT from Llama) 
& 0
& \textbf{\mstd{0.298}{0.145}} & \mstd{0.276}{0.118}
& \textbf{\mstd{0.236}{0.334}} & \textbf{\mstd{0.252}{0.293}}
& \textbf{\mstd{0.301}{0.195}} & \textbf{\mstd{0.285}{0.165}}
& \textbf{\mstd{0.269}{0.260}} & \textbf{\mstd{0.267}{0.224}} \\

AO 
& Qwen3-4B 
& 1
& \mstd{0.292}{0.135} & \textbf{\mstd{0.279}{0.109}}
& \mstd{0.131}{0.197} & \mstd{0.187}{0.188}
& \mstd{0.259}{0.157} & \mstd{0.256}{0.123}
& \mstd{0.210}{0.186} & \mstd{0.231}{0.159} \\

LatentQA 
& Qwen3-4B 
& 1
& \mstd{0.284}{0.140} & \mstd{0.266}{0.108}
& \mstd{0.199}{0.312} & \mstd{0.225}{0.278}
& \mstd{0.290}{0.176} & \mstd{0.275}{0.144}
& \mstd{0.246}{0.245} & \mstd{0.249}{0.211} \\

SelfIE 
& Qwen3-4B 
& 3
& \mstd{0.122}{0.056} & \mstd{0.187}{0.053}
& \mstd{0.020}{0.038} & \mstd{0.053}{0.047}
& \mstd{0.117}{0.065} & \mstd{0.176}{0.060}
& \mstd{0.073}{0.071} & \mstd{0.122}{0.083} \\

PatchScope 
& Qwen3-4B 
& 1
& \mstd{0.126}{0.059} & \mstd{0.190}{0.054}
& \mstd{0.019}{0.038} & \mstd{0.055}{0.047}
& \mstd{0.120}{0.066} & \mstd{0.176}{0.062}
& \mstd{0.075}{0.074} & \mstd{0.124}{0.084} \\

\midrule
\multicolumn{11}{l}{\textit{Donor: Llama-3.1-8B-Instruct}} \\
\rowcolor{lightblue} UAV 
& Llama-8B (Full) 
& 0
& \mstd{0.307}{0.147} & \mstd{0.283}{0.120}
& \textbf{\mstd{0.294}{0.368}} & \textbf{\mstd{0.303}{0.334}}
& \mstd{0.288}{0.175} & \mstd{0.271}{0.144}
& \textbf{\mstd{0.297}{0.277}} & \textbf{\mstd{0.290}{0.247}} \\

UAV 
& Qwen3-4B (Full) 
& 0
& \mstd{0.309}{0.150} & \mstd{0.284}{0.124}
& \mstd{0.262}{0.349} & \mstd{0.281}{0.313}
& \mstd{0.293}{0.182} & \mstd{0.277}{0.155}
& \mstd{0.284}{0.267} & \mstd{0.281}{0.236} \\

UAV 
& Qwen3-4B (AOT from Qwen) 
& 0
& \mstd{0.298}{0.139} & \mstd{0.273}{0.113}
& \mstd{0.239}{0.335} & \mstd{0.257}{0.299}
& \mstd{0.285}{0.180} & \mstd{0.275}{0.158}
& \mstd{0.268}{0.257} & \mstd{0.266}{0.226} \\

AO 
& Llama-8B 
& 1
& \mstd{0.311}{0.145} & \textbf{\mstd{0.285}{0.120}}
& \mstd{0.262}{0.351} & \mstd{0.277}{0.318}
& \mstd{0.293}{0.185} & \mstd{0.275}{0.154}
& \mstd{0.285}{0.268} & \mstd{0.279}{0.238} \\

LatentQA 
& Llama-8B 
& 1
& \textbf{\mstd{0.313}{0.150}} & \textbf{\mstd{0.285}{0.124}}
& \mstd{0.280}{0.367} & \mstd{0.296}{0.334}
& \textbf{\mstd{0.306}{0.182}} & \textbf{\mstd{0.280}{0.151}}
& \mstd{0.296}{0.278} & \mstd{0.289}{0.249} \\

SelfIE 
& Llama-8B 
& 3
& \mstd{0.123}{0.081} & \mstd{0.164}{0.062}
& \mstd{0.022}{0.053} & \mstd{0.074}{0.063}
& \mstd{0.128}{0.109} & \mstd{0.176}{0.088}
& \mstd{0.077}{0.092} & \mstd{0.124}{0.083} \\

PatchScope 
& Llama-8B 
& 1
& \mstd{0.156}{0.080} & \mstd{0.192}{0.058}
& \mstd{0.018}{0.039} & \mstd{0.066}{0.050}
& \mstd{0.138}{0.099} & \mstd{0.182}{0.078}
& \mstd{0.088}{0.096} & \mstd{0.131}{0.085} \\

\bottomrule
\end{tabular}}
\vspace{2pt}

\begin{minipage}{0.98\textwidth}
\footnotesize
\textit{Note.}
All entries report the mean and standard deviation across individual evaluation examples. The standard deviation reflects per-example variation in open-ended generation rather than variation across random seeds. \textit{Full} denotes full two-stage adaptation on the current donor–decoder pair: Stage 1 trains the adapter with the decoder frozen, and Stage 2 jointly trains the adapter and decoder-side LoRA while keeping the decoder backbone frozen. AOT denotes adapter-only transfer. In both AOT settings, the current decoder is Qwen3-4B, and ``AOT from X'' indicates that the frozen decoder-side LoRA was learned through Full adaptation using donor X. For the Qwen3-4B donor with AOT from Llama, we reuse the LoRA learned with the Llama-8B donor and Qwen3-4B decoder. For the Llama-8B donor with AOT from Qwen, we reuse the LoRA learned with the Qwen3-4B donor and Qwen3-4B decoder. In each case, only a new adapter for the current donor is trained.
\end{minipage}
\end{table*}
\paragraph{ROUGE-L.}
ROUGE-L measures sequence-level similarity based on the longest common subsequence (LCS) between the generated answer and the reference~\cite{lin2004rouge}. 
Given a generated sequence $x$ and a reference sequence $y$, let $\mathrm{LCS}(x,y)$ denote the length of their longest common subsequence. 
The LCS-based recall and precision are:
\begin{equation}\label{eq:lcs}
R_{\mathrm{LCS}} = \frac{\mathrm{LCS}(x,y)}{|y|}, 
\quad
P_{\mathrm{LCS}} = \frac{\mathrm{LCS}(x,y)}{|x|}.
\end{equation}
ROUGE-L is then computed as:
\begin{equation}\label{eq:rougel}
\mathrm{ROUGE\text{-}L} =
\frac{(1+\beta^2) R_{\mathrm{LCS}} P_{\mathrm{LCS}}}
{R_{\mathrm{LCS}} + \beta^2 P_{\mathrm{LCS}}},
\end{equation}
where $\beta$ controls the relative importance of recall and precision.

\paragraph{chrF++.}
chrF++ computes an $F$-score over character $n$-grams and word $n$-grams~\cite{popovic-2017-chrf}. 
Let $P_{\mathrm{char}}$ and $R_{\mathrm{char}}$ denote the average precision and recall over character $n$-grams, and let $P_{\mathrm{word}}$ and $R_{\mathrm{word}}$ denote the corresponding averages over word $n$-grams. 
The combined precision and recall are:
\begin{equation}\label{eq:char-word}
P = \frac{P_{\mathrm{char}} + P_{\mathrm{word}}}{2},
\ 
R = \frac{R_{\mathrm{char}} + R_{\mathrm{word}}}{2}.
\end{equation}
chrF++ is then computed as:
\begin{equation}\label{eq:chrf}
\mathrm{chrF{++}} =
\frac{(1+\beta^2)PR}{\beta^2 P + R},
\end{equation}
where $\beta$ controls the relative weight of recall and precision. 
Following the standard setting, we use character $n$-grams up to order 6, word $n$-grams up to order 2, and $\beta=2$.

\paragraph{BERTScore.}
BERTScore measures semantic similarity by comparing contextualized token embeddings of the generated answer and the reference~\cite{zhang2019bertscore}. 
Given generated tokens $x=\{x_i\}_{i=1}^{m}$ and reference tokens $y=\{y_j\}_{j=1}^{n}$, BERTScore first computes pairwise cosine similarities between their embeddings. 
Precision and recall are then defined by greedy token matching:
\begin{equation}
\begin{aligned}
P_{\mathrm{BERT}} 
&= \frac{1}{m}\sum_{i=1}^{m}\max_{j} \cos(\mathbf{x}_i, \mathbf{y}_j), \\
R_{\mathrm{BERT}} 
&= \frac{1}{n}\sum_{j=1}^{n}\max_{i} \cos(\mathbf{x}_i, \mathbf{y}_j).
\end{aligned}
\end{equation}
The final score is their F1:
\begin{equation}
F_{\mathrm{BERT}} =
\frac{2 P_{\mathrm{BERT}} R_{\mathrm{BERT}}}
{P_{\mathrm{BERT}} + R_{\mathrm{BERT}}}.
\end{equation}
Because it compares contextualized embeddings rather than exact tokens, BERTScore can reward semantically similar paraphrases with different surface forms.

\subsection{Source-Grounded LLM-as-a-Judge Evaluation}
\label{app:llm_judge}

We randomly sample 50 examples from each task and evaluate them using
GPT-4.1-mini and Qwen2.5-14B. Each judge is given the source text,
question, and model prediction, and separately evaluates topic relevance
and detail-level factual consistency on a scale from 0 to 5. A score of
at least 3 indicates that the prediction is at least partially supported
by the source.

\begin{table*}[t]
    \centering
    \small
    \begin{tabular}{llcccccc}
        \toprule
        \textbf{Task} & \textbf{Judge}
        & \textbf{R-L} & \textbf{BERTS}
        & \textbf{Topic} & \textbf{Topic $\geq 3$}
        & \textbf{Detail} & \textbf{Detail $\geq 3$} \\
        \midrule

        Fact retrieval
        & GPT-4.1-mini
        & 0.233 & 0.374 & 3.60 & 70\% & 1.78 & 24\% \\

        & Qwen2.5-14B
        & 0.233 & 0.374 & 3.36 & 64\% & 1.90 & 30\% \\

        \midrule

        Comprehension
        & GPT-4.1-mini
        & 0.297 & 0.343 & 3.06 & 54\% & 1.76 & 24\% \\

        & Qwen2.5-14B
        & 0.297 & 0.343 & 2.56 & 40\% & 1.44 & 26\% \\

        \midrule

        Gist
        & GPT-4.1-mini
        & 0.204 & 0.247 & 1.34 & 16\% & 0.62 & 6\% \\

        & Qwen2.5-14B
        & 0.204 & 0.247 & 1.42 & 8\% & 0.38 & 4\% \\

        \bottomrule
    \end{tabular}
    \caption{
        Source-grounded LLM-as-a-judge evaluation. R-L and BERTS are
        computed on the full test set, while each judge evaluates 50
        randomly sampled examples per task. Topic and Detail report the
        mean scores on a 0--5 scale.
    }
    \label{tab:llm_judge}
\end{table*}

As shown in Table~\ref{tab:llm_judge}, UAV recovers coarse topic-level
information more reliably than fine-grained factual details. For fact
retrieval, 64--70\% of predictions receive a topic score of at least 3,
whereas only 24--30\% receive a detail score of at least 3.
Comprehension shows intermediate performance, while gist remains the
most difficult task, with only 8--16\% reaching the topic threshold and
4--6\% reaching the detail threshold. These results suggest that UAV can
recover nontrivial coarse topic-level signals, but fine-grained factual
details and abstract gist remain unreliable.

\paragraph{Qualitative failure case.}
The following example illustrates how reference-based metrics may reward
shared words or entities despite factual errors.

\begin{quote}
\textbf{Source:} Syria Redeploys Some Security Forces in Lebanon.
Syria, under intense pressure to quit Lebanon, pulled out its security
forces from several locations and redeployed them elsewhere.

\textbf{Question:} What concept is this text mainly about?

\textbf{Prediction:} Lebanese security forces are moving into a refugee
camp in the south to prevent an influx of Palestinian refugees.

\textbf{Reference:} Syria withdrew its security forces from key
locations in Beirut and northern Lebanon and redeployed them in eastern
Lebanon amid pressure to leave the country.
\end{quote}

The prediction remains broadly related to security forces and Lebanon
but changes the actor, action, and event. It may therefore receive
partial topic credit while remaining factually incorrect at the detail
level.

\subsection{Paired-Bootstrap Significance Tests}
\label{app:significance}

We conduct paired-bootstrap tests over the same 1,500 evaluation
examples used in the main comparison. We report the mean paired
difference, its 95\% confidence interval, and the example-level win
rate. A positive difference favors the first method in each comparison.

\begin{table*}[t]
    \centering
    \small
    \begin{tabular}{llccc}
        \toprule
        \textbf{Comparison} &
        \textbf{Metric} &
        \textbf{$\Delta$} &
        \textbf{95\% CI} &
        \textbf{Win Rate} \\
        \midrule

        UAV vs.\ AO (Qwen3-4B)
        & ROUGE-L
        & $+0.056$
        & $[0.045, 0.066]$
        & 59.6\% \\

        & BERTScore
        & $+0.110$
        & $[0.097, 0.124]$
        & 63.6\% \\

        \midrule

        UAV vs.\ LatentQA (Qwen3-4B)
        & ROUGE-L
        & $+0.019$
        & $[0.009, 0.028]$
        & 54.0\% \\

        & BERTScore
        & $+0.017$
        & $[0.007, 0.026]$
        & 52.9\% \\

        \midrule

        Self vs.\ Cross (Llama-3.1-8B)
        & ROUGE-L
        & $+0.012$
        & $[0.002, 0.022]$
        & 53.5\% \\

        & BERTScore
        & $+0.010$
        & $[-0.000, 0.020]$
        & 51.5\% \\

        \midrule

        Self vs.\ Cross (Gemma-3-4B)
        & ROUGE-L
        & $+0.004$
        & $[-0.006, 0.015]$
        & 49.4\% \\

        & BERTScore
        & $-0.004$
        & $[-0.014, 0.006]$
        & 48.6\% \\

        \bottomrule
    \end{tabular}
    \caption{
        Paired-bootstrap comparisons over 1,500 aligned evaluation
        examples. Positive differences favor the first method in each
        comparison. The win rate is the percentage of paired examples
        on which the first method obtains a higher score.
    }
    \label{tab:significance}
\end{table*}

The confidence intervals for UAV versus AO and LatentQA exclude zero.
The improvements over AO are relatively large, whereas those over
LatentQA are statistically detectable but small. For Llama-3.1-8B,
ROUGE-L shows a small self-decoding advantage, while the BERTScore
interval reaches approximately zero. For Gemma-3-4B, both confidence
intervals include zero, and the directions differ across metrics.
Therefore, we characterize self- and cross-decoding as largely
comparable rather than claiming that either setting consistently
outperforms the other.

These confidence intervals measure uncertainty in paired performance
differences. They are distinct from the standard deviations in the main
tables, which represent variation across individual evaluation examples,
and from the multi-seed results, which measure training stability.

\section{Additional Experimental Results}\label{app:exp}

\subsection{Training Efficiency}
\label{app:efficiency}

We compare the training cost of 12B self-decoding with UAV
cross-decoding using a 4B external decoder. GPU-hours are measured per
epoch under the same experimental setup.

\begin{table}[t]
    \centering
    \small
    \resizebox{\columnwidth}{!}{
    \begin{tabular}{lccc}
        \toprule
        \textbf{Setting}
        & \textbf{Stage 1}
        & \textbf{Stage 2}
        & \textbf{Best Val. Loss} \\
        & \textbf{GPU-h/epoch}
        & \textbf{GPU-h/epoch}
        & \\
        \midrule

        Self-decoding (12B)
        & 7.1
        & 26.5
        & \textbf{1.6004} \\

        UAV cross-decoding (4B)
        & \textbf{2.7}
        & \textbf{11.6}
        & 1.6594 \\

        \bottomrule
    \end{tabular}
    }
    \caption{
        Training efficiency of 12B self-decoding and UAV
        cross-decoding with a 4B external decoder. GPU-hours are
        reported per epoch.
    }
    \label{tab:efficiency}
\end{table}

As shown in Table~\ref{tab:efficiency}, UAV reduces the Stage-1 cost
from 7.1 to 2.7 GPU-hours per epoch and the Stage-2 cost from 26.5 to
11.6 GPU-hours per epoch. The best validation loss increases from
1.6004 to 1.6594, corresponding to a difference of 0.059.

The efficiency gain comes from caching donor activations before
training. UAV therefore does not need to keep the donor model in GPU
memory or backpropagate through it during verbalizer training. This
allows a smaller shared decoder to analyze activations from a larger
donor model.

\subsection{Training Stability Across Random Seeds}
\label{app:seed}

We evaluate training stability using five Stage-2 runs under the
500K-QA setting. The random seed affects both data shuffling and
parameter initialization. As shown in Table~\ref{tab:multiseed},
the small standard deviations indicate that Stage-2 training is
stable across random seeds.

\begin{table}[t]
    \centering
    \caption{Training stability across five random seeds under the 500K-QA Stage-2 setting. The standard deviation is computed across training runs.}\label{tab:multiseed}
    \small
    \begin{tabular}{ccc}
        \toprule
        \textbf{Train Seed} & \textbf{Validation Loss} & \textbf{PPL} \\
        \midrule
        42 & 1.8430 & 6.316 \\
        43 & 1.8414 & 6.306 \\
        44 & 1.8437 & 6.320 \\
        45 & 1.8443 & 6.323 \\
        46 & 1.8418 & 6.308 \\
        \midrule
        \textbf{Mean $\boldsymbol{\pm}$ Std.}
        & $\mathbf{1.8428 \pm 0.0012}$
        & $\mathbf{6.314 \pm 0.007}$ \\
        \bottomrule
    \end{tabular}
\end{table}

\subsection{Additional Adapter Architecture Results}\label{app:architecture}
Table~\ref{tab:app-adapter-arch} reports additional adapter architecture ablations with token-level and character-level generation metrics. 
For MLP-style adapters, increasing the bottleneck size generally improves chrF++, with the no-bottleneck variant achieving the highest chrF++ score among MLP adapters. 
However, the trend is less consistent for Tok-F1, suggesting that simply increasing the MLP capacity does not always lead to uniformly better lexical alignment. 
A similar pattern appears when varying the number of projected soft tokens: using more soft tokens does not monotonically improve performance, and the largest setting with 128 tokens even leads to lower scores, despite using more than one billion trainable adapter parameters. 
This indicates that over-parameterized MLP projections may introduce redundancy or optimization difficulty rather than consistently improving activation-to-text alignment.

\begin{table*}[t]
    \centering
    \caption{
        Dataset-level performance of UAV under the Qwen3-4B
        self-decoding setting. Each dataset contains 100 evaluation
        examples. R-L denotes ROUGE-L, and BERTS denotes BERTScore.
        The best result in each column is highlighted in bold.
    }
    \label{tab:dataset_level_results}
    \begin{tabular}{llcc}
        \toprule
        \textbf{Task Family} &
        \textbf{Dataset} &
        \textbf{R-L} &
        \textbf{BERTS} \\
        \midrule

        Concrete Wikipedia entities
        & \texttt{wikipedia\_person}
        & \textbf{0.319} & \textbf{0.475} \\

        & \texttt{wikipedia\_place}
        & 0.266 & 0.450 \\

        & \texttt{wikipedia\_work}
        & 0.189 & 0.388 \\

        & \texttt{wikipedia\_event}
        & 0.275 & 0.372 \\

        & \texttt{wikipedia\_organization}
        & 0.197 & 0.341 \\

        \midrule

        Abstract/general Wikipedia knowledge
        & \texttt{wikipedia\_concept}
        & 0.202 & 0.313 \\

        & \texttt{wikipedia\_generic}
        & 0.180 & 0.282 \\

        \midrule

        Topic and general-language tasks
        & \texttt{ag\_news}
        & 0.260 & 0.378 \\

        & \texttt{scientific}
        & 0.243 & 0.350 \\

        & \texttt{lmsys\_user}
        & 0.318 & 0.328 \\

        & \texttt{latentqa\_control}
        & 0.259 & 0.292 \\

        \midrule

        Sentiment and emotion tasks
        & \texttt{tweeteval\_emotion}
        & 0.297 & 0.319 \\

        & \texttt{sst2}
        & 0.260 & 0.307 \\

        & \texttt{dair\_emotion}
        & 0.302 & 0.297 \\

        & \texttt{tweeteval\_sentiment}
        & 0.241 & 0.274 \\

        \bottomrule
    \end{tabular}
\end{table*}

For Q-Former adapters, increasing depth and context width provides moderate gains over the shallow baseline, but the most effective configuration is obtained by increasing the decoder-side LoRA capacity. 
Specifically, QF-LoRA achieves the best Tok-F1 and chrF++ scores while using substantially fewer adapter parameters than the largest MLP variants. 
This suggests that the attention-based adapter is more parameter-efficient for transforming donor activations into decoder-readable representations, and that sufficient decoder-side adaptation is important for exploiting these representations. 
Based on these results, we adopt the Q-Former adapter with enhanced LoRA capacity as our default architecture.

\subsection{Dataset-Level and Error Analysis}
\label{app:dataset_error_analysis}

We conduct a dataset-level analysis using the Qwen3-4B
self-decoding setting from the main experiments. For each dataset,
we evaluate 100 examples and report ROUGE-L and BERTScore in
Table~\ref{tab:dataset_level_results}.

UAV performs best on concrete entity-centered retrieval tasks,
particularly \texttt{wikipedia\_person} and
\texttt{wikipedia\_place}, which achieve BERTScores of 0.475 and
0.450, respectively. Performance is lower on abstract or general
Wikipedia knowledge, with \texttt{wikipedia\_generic} obtaining a
BERTScore of 0.282. Sentiment and emotion datasets also show
relatively low BERTScores, ranging from 0.274 to 0.319. Results on
topic and general-language datasets are mixed, suggesting that
performance depends on the specificity and semantic structure of
the target information.

One possible explanation is that concrete entities are represented
more consistently in the model's pretraining distribution, whereas
abstract, affective, and semantically flexible tasks require more
context-dependent interpretation. This explanation is interpretive
rather than directly established by our experiments.

Our qualitative error analysis identifies two common failure modes.
First, when activation alignment is weak, UAV may produce fluent but
input-irrelevant content. Second, when fine-grained information is
not fully preserved in the donor activation, UAV may confuse
semantically similar entities. These cases suggest that generation
fluency does not necessarily imply faithful recovery of the
information encoded in the donor activation.

\subsection{Additional Comparison with Activation Interpretation Baselines}
\label{app:app-baseline}

Table~\ref{tab:main_results_token_chrf} provides complementary token-level and character-level results for the baseline comparison in Section~\ref{sec:baselines}. 
While the main paper reports ROUGE-L and BERTScore as high-level generation quality metrics, here we additionally report Token-F1 and chrF to evaluate lexical and character-level agreement with the reference answers.

The results are consistent with the main comparison in Table~\ref{tab:main_results}. 
For Qwen3-4B activations, UAV achieves the strongest overall performance under both full two-stage adaptation and adapter-only transfer. 
In particular, \textsc{AOT} from Llama slightly improves over the corresponding full self-explanation setting on both Token-F1 and chrF, showing that the frozen decoder-side LoRA learned from another donor can still provide an effective explanation-oriented decoder space. 
Compared with training-based self-explanation baselines, UAV obtains clear gains on fact retrieval, where recovering precise information from the activation is especially important.

\begin{figure*}[!ht]
    \centering
    \subfloat[Classification]{\includegraphics[width=0.32\linewidth]{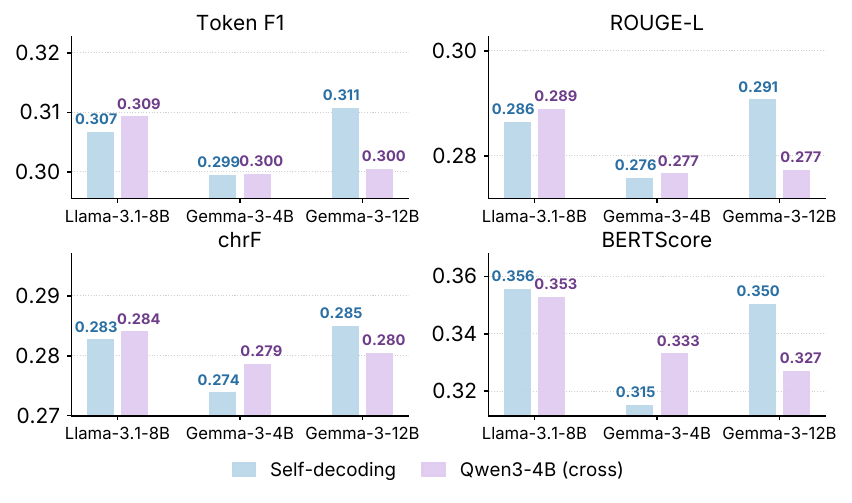}}
    \subfloat[Fact]{\includegraphics[width=0.32\linewidth]{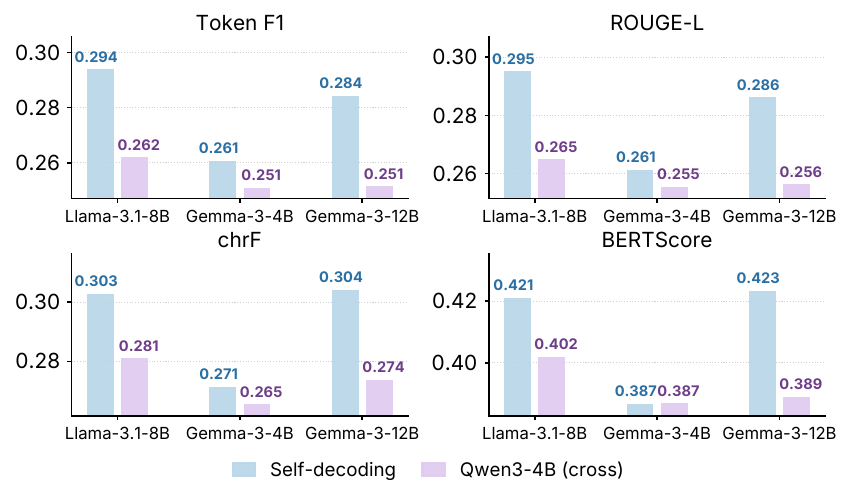}}
    \subfloat[Gist]{\includegraphics[width=0.32\linewidth]{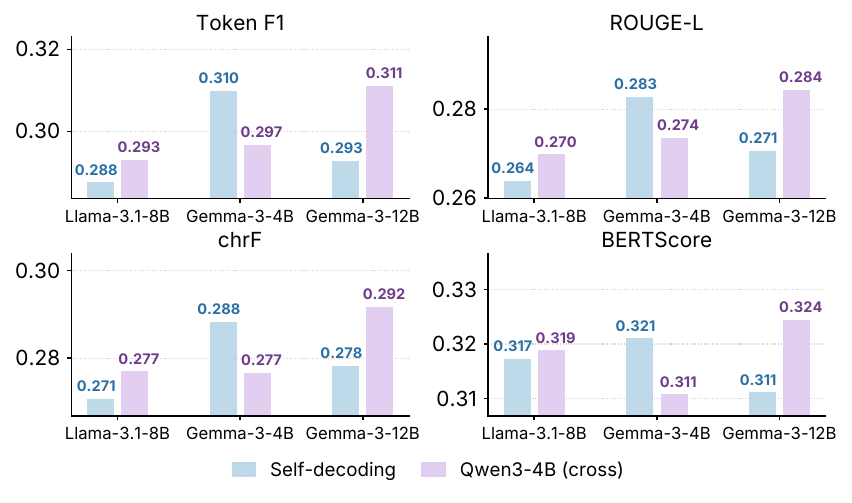}}
    \caption{Task-level comparison between self-decoding and cross-model verbalization on classification, fact retrieval, and gist summarization.}
    \label{fig:app:external}
\end{figure*}

For Llama-3.1-8B activations, the full Llama-8B verbalizer achieves the best overall Token-F1 and chrF among UAV variants and remains competitive with LatentQA. 
When using Qwen3-4B as an external decoder, UAV still maintains strong performance, demonstrating that the proposed adapter-decoder framework can verbalize activations across different model families rather than relying only on self-explanation. 
The adapter-only transfer setting incurs only a moderate drop from full adaptation while still substantially outperforming training-free baselines.

Across both donors, SelfIE and PatchScope perform much worse than trained verbalizers, especially on fact retrieval. 
This further supports the observation that directly patching activations into prompts is insufficient for open-ended activation-grounded QA. 
In contrast, UAV learns an explicit adapter that maps donor activations into decoder-readable soft tokens, leading to stronger lexical, character-level, and semantic agreement with the reference answers.

\subsection{Additional Results on Cross Model Decoding}~\label{app:cross-dec}
\begin{table}[t]
\centering
\setlength{\tabcolsep}{5pt}
\caption{
Additional cross-donor verbalization results with Qwen3-4B-Instruct-2507 as the shared decoder.
We report Token-F1 and chrF as complementary lexical and character-level metrics.
}
\label{tab:app:cross_dec}
\scalebox{0.86}{
\begin{tabular}{lcc}
\toprule
\textbf{Donor}
& \textbf{Tok-F1 $\uparrow$}
& \textbf{chrF $\uparrow$} \\
\midrule
Llama-3.1-8B-Instruct
& $0.284_{\pm 0.267}$
& $0.281_{\pm 0.236}$ \\

Gemma-3-4B-IT
& $0.276_{\pm 0.264}$
& $0.272_{\pm 0.231}$ \\

Gemma-3-12B-IT
& $0.280_{\pm 0.261}$
& $0.280_{\pm 0.230}$ \\

\rowcolor{lightblue}
Yi-1.5-34B-Chat
& $\mathbf{0.310}_{\pm 0.278}$
& $\mathbf{0.306}_{\pm 0.250}$ \\
\bottomrule
\end{tabular}
}
\begin{flushleft}
\footnotesize
\textit{Note.}
All rows use Qwen3-4B-Instruct-2507 as the shared decoder.
The best result in each column is highlighted in bold.
\end{flushleft}
\end{table}
Table~\ref{tab:app:cross_dec} reports complementary Token-F1 and chrF results for the cross-donor setting. 
The trends are consistent with the main-paper metrics in Table~\ref{tab:app:cross_dec}. 
Yi-1.5-34B achieves the strongest lexical and character-level agreement, obtaining the best Token-F1 and chrF among all donor models. 
Gemma-3-12B also improves over Gemma-3-4B, further suggesting that larger models within the same family can expose representations that are easier for the verbalizer to decode. 
However, the comparison across families again shows that scale alone does not fully determine decoding quality: Llama-3.1-8B remains stronger than both Gemma models despite being smaller than Gemma-3-12B.

Together with the main results, these additional metrics show that the cross-donor behavior is stable across semantic, lexical, and character-level evaluation. 
UAV can adapt the same Qwen3-4B decoder to activations from multiple donor families, while the remaining performance variation reflects both donor scale and family-specific representation structure.

\subsection{Task-level comparison between self-decoding and external-decoder verbalization}\label{app:external}
Figure~\ref{fig:app:external} provides task-level results for classification, fact retrieval, and gist summarization. 
The overall trend is consistent with the main-paper comparison: self-decoding usually performs better than cross-model verbalization, indicating that a family-matched decoder benefits from a more compatible activation space. 
However, the size of the gap varies across task types and donor models.

\begin{figure*}[!ht]
    \centering
    \includegraphics[width=0.23\linewidth]{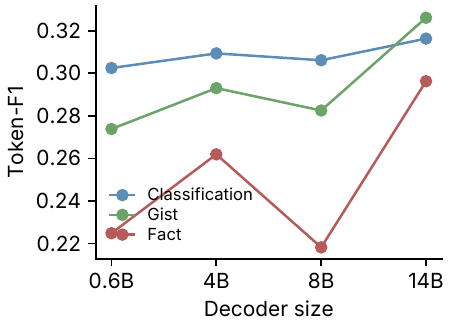}\hfill
    \includegraphics[width=0.23\linewidth]{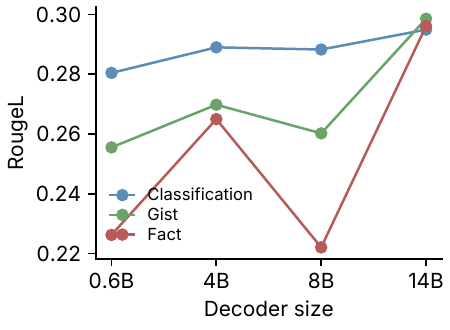}\hfill
    \includegraphics[width=0.23\linewidth]{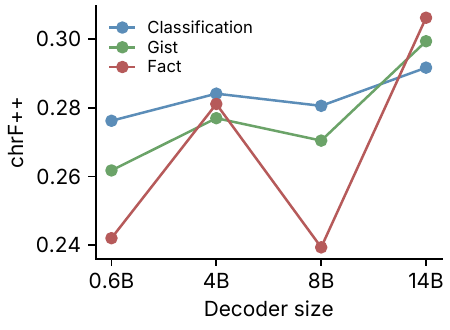}\hfill
    \includegraphics[width=0.23\linewidth]{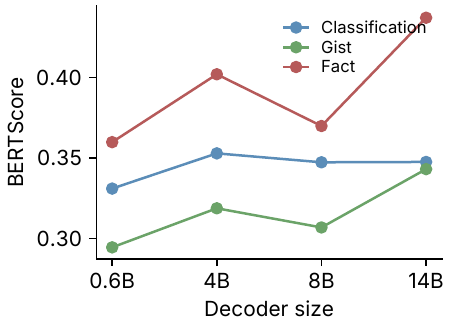}
    \caption{Task-level effect of decoder size on cross-decoding performance.
All results use Llama-3.1-8B-Instruct layer-27 activations as the donor and vary the Qwen-family decoder size.}
    \label{fig:app:decoder}
\end{figure*}

The difference is most consistent on fact retrieval. 
Across the fact-oriented subset, self-decoding almost always outperforms the Qwen3-4B cross-model decoder, especially for Llama-3.1-8B and Gemma-3-12B. 
This suggests that fine-grained factual recovery is more sensitive to representation-space mismatch than classification or gist summarization. 
Since fact retrieval requires the verbalizer to recover specific entity-level or attribute-level information from the activation, even small alignment errors between the donor activation space and the shared decoder embedding space can lead to degraded lexical and semantic agreement.

\begin{figure}[t]
    \centering
    \includegraphics[width=\linewidth]{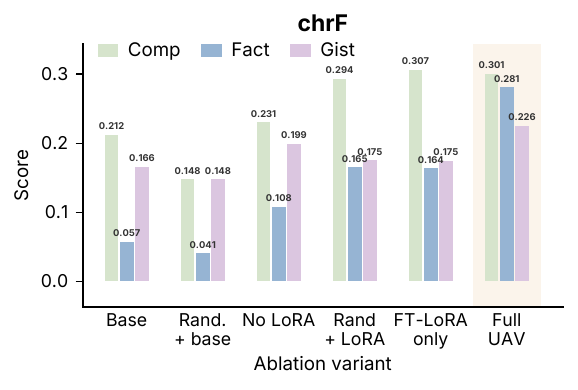}
    \caption{Additional chrF results for disentangling the roles of the activation adapter and decoder-side LoRA.}
    \label{fig:app:activation}
\end{figure}

For classification and gist summarization, the comparison is more mixed. 
Classification relies more on coarse label-level signals, while gist summarization mainly requires topic-level semantic information. 
These types of information appear more robust under cross-model transfer, and in several cases the Qwen3-4B cross-model decoder matches or slightly exceeds self-decoding on individual metrics. 
For example, cross-model verbalization improves BERTScore for Gemma-3-4B in classification and improves several gist metrics for Gemma-family donors. 
These results suggest that cross-model verbalization is more effective when the target query depends on high-level semantic information, but remains more challenging for precise factual extraction.

\begin{figure*}[!ht]
    \centering
    \includegraphics[width=0.19\linewidth]{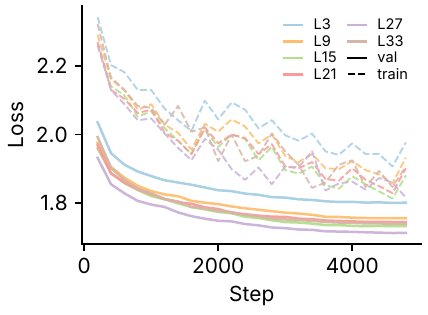}\hfill
    \includegraphics[width=0.19\linewidth]{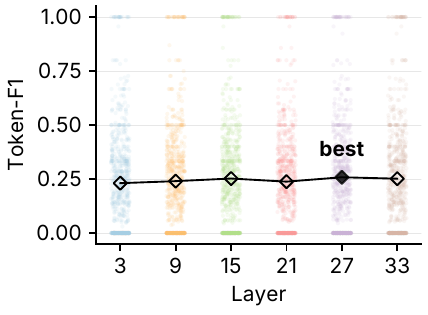}\hfill
    \includegraphics[width=0.19\linewidth]{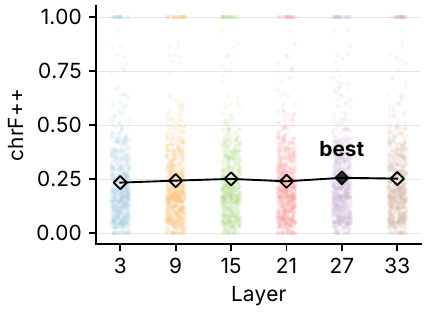}\hfill
    \includegraphics[width=0.19\linewidth]{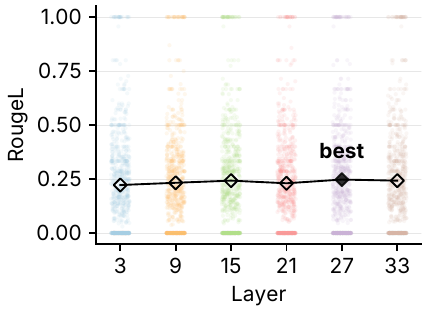}\hfill
    \includegraphics[width=0.19\linewidth]{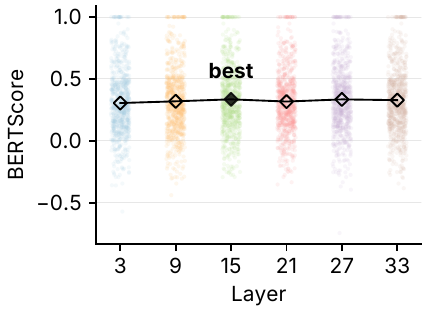}
    \caption{Additional layer-wise results under the self-explanation setting. 
We report training dynamics and evaluation results for Qwen3-4B activations from Layers 3, 9, 15, 21, 27, and 33. 
The verbalizer can learn from all evaluated layers, while middle-to-late layers generally yield stronger decoding performance.}
    \label{fig:app-layer}
\end{figure*}
\begin{figure*}[!ht]
    \centering
    \includegraphics[width=0.19\linewidth]{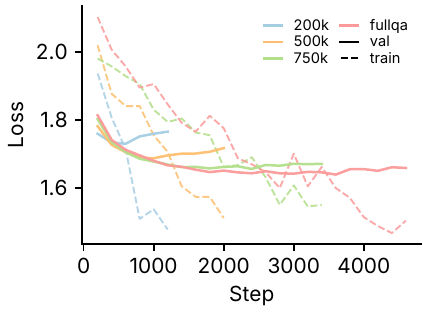}\hfill
    \includegraphics[width=0.19\linewidth]{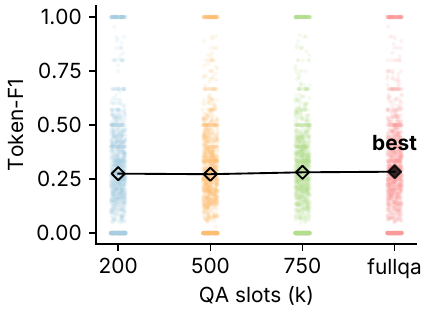}\hfill
    \includegraphics[width=0.19\linewidth]{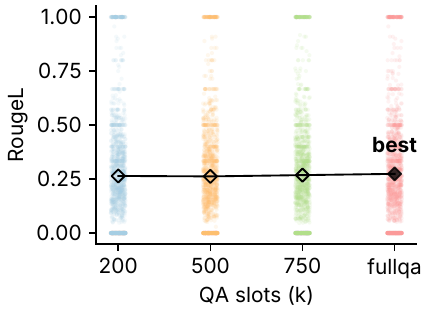}\hfill
    \includegraphics[width=0.19\linewidth]{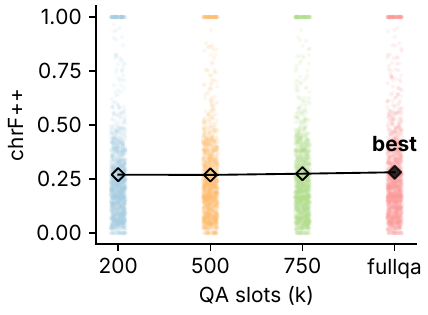}\hfill
    \includegraphics[width=0.19\linewidth]{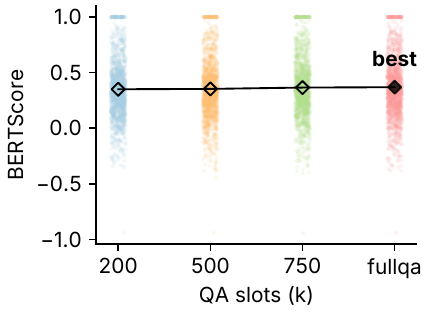}
    \caption{Additional training data scale results. 
We compare verbalizers trained with 200K, 422K, 750K, and 951K Stage-2 QA pairs. 
Training remains stable across all scales, and evaluation performance generally improves as the amount of query-oriented supervision increases. 
The full 951K setting achieves the strongest overall performance across token-level and semantic metrics.}
    \label{fig:app-datascale}
\end{figure*}

Together, these task-level results clarify the trade-off observed in the main paper. 
Self-decoding provides the strongest performance when model-specific decoder adaptation is available, particularly for fact retrieval. 
Cross-model verbalization, however, offers a more universal setup by using a shared decoder across different donor families, while preserving competitive performance on coarse-grained and semantic explanation tasks.

\subsection{Task-level Performance across Decoder sizes}\label{app:decoder}
Figure~\ref{fig:app:decoder} provides task-level results for the decoder-size study in Section~4.4.
All settings use Llama-3.1-8B-Instruct layer-27 activations as the donor and vary the Qwen-family decoder size.
The task-level results are broadly consistent with the overall results in the main paper: increasing the decoder
from 0.6B to 4B improves most metrics, while Qwen3-8B does not consistently outperform Qwen3-4B despite
having comparable validation loss. Qwen3-14B achieves the strongest performance in most task-level settings,
especially on fact retrieval and gist summarization.

The effect of decoder scaling is not uniform across tasks. Classification is relatively stable across decoder sizes,
suggesting that coarse label-level information is easier to recover once the activation has been mapped into the
decoder space. In contrast, fact retrieval is more sensitive to decoder capacity, as it requires recovering
fine-grained input-specific information from the donor activation. Gist summarization also benefits from the
larger decoder, indicating that stronger decoders can better convert aligned activation signals into high-level
semantic descriptions. These results suggest that decoder capacity helps activation verbalization, but scaling
alone does not guarantee monotonic improvements; decoder-specific optimization and instruction-following
behavior also affect final generation quality.
\subsection{More Results on Disentangling Adapter and Decoder Contributions}\label{app:activation}
Figure~\ref{fig:app:activation} provides additional chrF results for the ablation study in
Section~4.5, which disentangles the roles of the activation adapter and decoder-side LoRA.
The trend is consistent with the Token-F1, ROUGE-L, and BERTScore results in Figure~\ref{fig:activation}.
The base decoder and randomly initialized adapter variants perform poorly, showing that the model cannot
answer activation-grounded questions from the decoder prior or arbitrary soft-token injection alone.

Decoder-side LoRA improves classification-style performance, suggesting that Stage-2 tuning helps the decoder
learn the QA format and task-level decision patterns. However, LoRA-only variants remain much weaker on
fact retrieval and gist summarization, where the answer depends more heavily on input-specific information
encoded in the donor activation. In contrast, Full UAV obtains the strongest chrF scores on these
information-intensive tasks, further confirming that the trained adapter is necessary for extracting
activation-grounded factual and semantic evidence.

\begin{table*}[t]
\centering
\caption{Default optimization settings used by the main experiment and by
experiments marked as following the default recipe.  The global batch is the
nominal configured value.}
\label{tab:default-training}
\begin{tabular}{lcc}
\toprule
Setting & Stage 1 & Stage 2 \\
\midrule
Trainable modules & activation adapter & activation adapter + LoRA \\
Optimizer & AdamW & AdamW \\
Adapter learning rate & $3\times10^{-4}$ & $5\times10^{-5}$ \\
LoRA learning rate & -- & $1\times10^{-4}$ \\
Weight decay & 0.01 & 0.01 \\
Warmup ratio & 0.05 & 0.10 \\
Schedule & cosine to zero & cosine to zero \\
Maximum epochs & 16 & 10 \\
Per-rank microbatch & 128 & 128 \\
Gradient accumulation & 3 & 3 \\
Number of ranks & 4 & 4 \\
Nominal global batch & 1536 & 1536 \\
Gradient clipping & 1.0 & 1.0 \\
Evaluation interval & 200 updates & 200 updates \\
Early-stop patience & 3 & 5 \\
Minimum improvement & 0 & 0 \\
\bottomrule
\end{tabular}
\end{table*}

\subsection{Additional Layer-wise Results.}\label{app:layerwise}
We provide additional layer-wise results to complement the analysis in Section~\ref{sec:ablations}. 
Specifically, we evaluate Qwen3-4B activations from Layers 3, 9, 15, 21, 27, and 33 under the self-explanation setting. 
Figure~\ref{fig:app-layer} reports both the training dynamics and token-level evaluation results. 
The training and validation losses decrease steadily across all examined layers, suggesting that the verbalizer can learn meaningful activation-to-text mappings from different depths. 
Meanwhile, the token-level metrics show a mild but consistent advantage for middle-to-late layers, with Layer 27 achieving the best Token-F1 and chrF++ scores among the evaluated layers. 
These additional results are consistent with the observation in the main text that later activations tend to be more informative for verbalization, while early-layer activations are comparatively less effective.

\subsection{Additional Training Data Scale Results}
\label{app:data_scale}
We provide additional results on the effect of Stage-2 training data scale in Figure~\ref{fig:app-datascale}. 
Specifically, we train the verbalizer with 200K, 422K, 750K, and 951K QA pairs and report both training dynamics and token-/semantic-level evaluation results in Figure~\ref{fig:app-datascale}. 
The training curves show stable optimization across different data scales. 
Meanwhile, the evaluation results indicate that increasing the number of QA pairs consistently improves verbalization quality, especially for the full 951K setting. 
These results further support the observation in Section~\ref{sec:ablations} that larger and more diverse query-oriented supervision helps the verbalizer better interpret activations under downstream queries.

%

\section{Training Details}
\label{app:training-details}

\subsection{Default two-stage recipe}

Unless otherwise noted, our experiments use the following two-stage training
procedure.  The default experiment takes the layer-27 residual-stream
activation of Llama-3.1-8B-Instruct as the donor representation and uses
Qwen3-4B as the decoder.  The donor activation is a single
$4096$-dimensional vector, while the decoder hidden size is $2560$.

\paragraph{Activation adapter.}
We map each donor activation to a sequence of 64 soft tokens using a
cross-attention adapter.  A linear projection first converts the donor vector
into eight decoder-dimensional context vectors.  Sixty-four learned query
vectors then attend to these contexts through two pre-normalized
cross-attention blocks.  Each block uses eight attention heads and a feed-forward
network with expansion factor four.  Dropout is 0.1 throughout the adapter.
The resulting adapter contains approximately 241M trainable parameters.

\paragraph{Stage 1: activation reconstruction.}
In Stage 1, the decoder language model is frozen and only the activation
adapter is optimized.  For each example, the adapter output replaces 64
activation-placeholder embeddings, followed by the fixed prompt
\texttt{What does this representation encode?}.  The objective is causal
language modeling of the source text.  Placeholder and prompt tokens are
masked from the loss, and reconstruction targets are truncated to at most 128
tokens.  We use AdamW with adapter learning rate $3\times10^{-4}$, weight decay
0.01, and gradient-norm clipping at 1.0.  The learning rate is linearly warmed
up for 5\% of the planned optimizer updates and then cosine-decayed to zero.
Training runs for at most 16 epochs, with validation every 200 optimizer
updates and early stopping after three consecutive non-improving evaluations.

\paragraph{Stage 2: question answering.}
In Stage 2, we initialize the activation adapter from the selected Stage-1
checkpoint and jointly optimize it with LoRA modules attached to the frozen
decoder.  The default LoRA configuration has rank 64, scaling parameter 128,
dropout 0.1, no bias parameters, and targets all linear modules.  We supervise
only the assistant answer in the decoder chat template; the user prompt and
activation placeholders are masked from the loss, and decoder thinking is
disabled.  We again use AdamW, with learning rates $1\times10^{-4}$ for LoRA
and $5\times10^{-5}$ for the activation adapter.  We use weight decay 0.01,
gradient-norm clipping at 1.0, 10\% linear warmup, and cosine decay to zero.
Training runs for at most 10 epochs, with validation every 200 optimizer
updates and early stopping after five non-improving evaluations.

Both stages use token-level causal cross-entropy.  AdamW uses the PyTorch
defaults $\beta_1=0.9$, $\beta_2=0.999$, and $\epsilon=10^{-8}$.  If $T$ is
the planned number of optimizer updates, $W$ is the number of warmup updates,
and $s$ is the current update, the default learning-rate multiplier is
\begin{equation}
\lambda(s) =
\begin{cases}
s / \max(1,W), & s < W,\\
\frac{1}{2}\left[1+\cos\left(\pi\frac{s-W}{\max(1,T-W)}\right)\right],
& s \ge W.
\end{cases}
\end{equation}
The scheduler is advanced after each optimizer update, rather than after each
microbatch.

\paragraph{Precision and distributed training.}
The base decoder and activation adapter are loaded in bfloat16, and activation
checkpointing is enabled in both stages.  PEFT is invoked with adapter-dtype
autocasting enabled, which promotes trainable LoRA weights to float32 while
retaining bfloat16 model computation.  The default
jobs use four A100 GPUs with one distributed process per GPU.  The Stage-1
decoder is frozen and replicated, while the activation adapter is wrapped in
distributed data parallelism.  In Stage 2, the LoRA-augmented decoder and the
activation adapter are wrapped separately in distributed data parallelism.
The configured per-rank microbatch size is 128 with three-step gradient
accumulation, corresponding to a nominal global update batch of 1536.

\subsection{Data splits and checkpoint selection}

The default Stage-1 corpus contains 469,309 raw records from 21 sources.  We
hold out 5\% for validation and assign 80\% of the remaining records to the
reconstruction objective, resulting in approximately 356,675 Stage-1 training
examples.  Although this split is called \texttt{test} in the implementation,
it is used for early stopping and checkpoint selection and is therefore a
validation split.

The clean-v2 Stage-2 corpus contains approximately 950,716 training QA slots
from 15 retained sources.  The main experiment uses all available training
slots.  Stage 2 uses physically separate training, validation, and test
directories; the test split is not accessed during training.  We select the
checkpoint with the lowest validation cross-entropy and promote it to the
final checkpoint.  All data splitting and data subsampling use seed 42.
Stage 2 additionally seeds Python, NumPy, and PyTorch with 42, with a
rank-specific CUDA random stream.  Stage 1 uses seed 42 for data splitting and
sampler order, but its adapter initialization and dropout random states were
not explicitly seeded.

\subsection{Deviations from the default recipe}

Most experiments retain the default recipe and change only the experimental
factor under study.  Table~\ref{tab:training-deviations} lists all material
exceptions among the reported experiments.  Exploratory legacy configurations
that are not used in reported comparisons are omitted.

\begin{table*}[t]
\centering
\small
\setlength{\tabcolsep}{5pt}
\renewcommand{\arraystretch}{1.08}
\caption{
Training settings for experiments that deviate from the default recipe in
Table~\ref{tab:default-training}. Unlisted settings retain their default
values. Microbatch/accumulation denotes the per-process microbatch size and
the number of gradient-accumulation steps, respectively. Stage-2 learning
rates are reported as LoRA/adapter.
}
\label{tab:training-deviations}
\begin{tabular}{
    p{0.19\textwidth}
    p{0.25\textwidth}
    p{0.50\textwidth}
}
\toprule
Experiment
& Experimental change
& Deviations from the default recipe \\
\midrule

\multicolumn{3}{l}{\textit{Data and training-stage ablations}} \\

Data scaling
& Use 200k, 422k, 750k, or 951k Stage-2 QA examples
& All variants initialize from the same Stage-1 checkpoint and retain the
  default optimization settings. Stage-2 data are subsampled at ratios
  0.2104, 0.4437, 0.7889, and 1.0, respectively. \\

\addlinespace[2pt]
No Stage-1 pretraining
& Remove activation-reconstruction pretraining
& Stage 2 follows the default recipe, but the activation adapter is randomly
  initialized rather than loaded from a Stage-1 checkpoint. \\

\midrule
\multicolumn{3}{l}{\textit{Model and representation scaling}} \\

Decoder scaling
& Use Qwen3 decoders of size 0.6B, 4B, 8B, 14B, or 32B
& The 0.6B, 4B, and 8B variants follow the default recipe. The 14B variant
  uses microbatch/accumulation 64/6. The 32B variant uses 128/3 in Stage 1
  and 4/96 in Stage 2. Stage-1 training for large decoders uses FSDP. \\

\addlinespace[2pt]
Cross-donor scaling
& Use Gemma-3 4B/12B or Yi-1.5 34B as the donor
& The Gemma variants follow the default optimization recipe. The Yi-1.5 34B
  Stage-1 run clips gradients at 0.5 instead of 1.0. \\

\addlinespace[2pt]
Self-decoding
& Use the same model as both donor and decoder
& The Llama-8B and Gemma variants use the default learning rates. For Gemma,
  LoRA is applied only to the language-model projection layers. Gemma-3 12B
  Stage 2 uses eight distributed processes with microbatch/accumulation 64/3,
  preserving an effective global batch size of 1536. \\

\addlinespace[2pt]
Donor-layer ablation
& Use Qwen3-4B donor layers 3, 9, 15, 21, 27, or 33
& All variants use the same ablation-specific recipe: Qwen3-4B self-decoding,
  LoRA rank/alpha 16/32, a reconstruction ratio of 1.0, Stage-1 and Stage-2
  adapter learning rates of $10^{-3}$ and $3\times10^{-4}$, respectively,
  eight Stage-2 epochs, and early-stopping patience of six epochs in both
  stages. \\

\midrule
\multicolumn{3}{l}{\textit{Adapter architecture ablations}} \\

MLP adapter sweep
& Replace cross-attention with an MLP and vary the number of soft tokens and
  bottleneck size
& The Stage-1 adapter learning rate is $10^{-3}$. Most bottleneck variants
  use a LoRA dropout of 0.05 and Stage-2 learning rates of
  $10^{-5}/10^{-4}$. The v0-4 variant instead uses
  $5\times10^{-5}/5\times10^{-4}$. Variants with more soft tokens use
  smaller microbatches while preserving the effective global batch size. \\

\addlinespace[2pt]
Cross-attention sweep
& Vary adapter depth (one or four blocks) and context count (eight or 16)
& The v1-1 variant uses a Stage-1 adapter learning rate of $10^{-3}$.
  The v1-1, v1-3, and v1-4 variants use LoRA rank/alpha 16/32 and a Stage-2
  adapter learning rate of $3\times10^{-4}$; v1-3 and v1-4 additionally use
  smaller microbatches. \\

\midrule
\multicolumn{3}{l}{\textit{Transfer and additional ablations}} \\

Frozen-decoder transfer
& Train an adapter for a new donor while retaining an existing Stage-2 decoder
& The decoder backbone and LoRA parameters are frozen, and only the new
  activation adapter is optimized. The adapter is trained with a learning
  rate of $5\times10^{-4}$ for at most 10 epochs. \\

\addlinespace[2pt]
Additional model and task ablations
& Compare base and instruction-tuned models and different task subsets
& Stage 2 follows the default recipe using the 0.4437 data subsample.
  Instruction-tuned self-decoding variants initialize from the existing
  Qwen layer-27 Stage-1 checkpoint, which uses the layer-ablation recipe.
  The continual gist/classification-to-factual experiment initializes from
  the preceding Stage-2 checkpoint but resets the optimizer and learning-rate
  scheduler. \\

\bottomrule
\end{tabular}
\end{table*}

\paragraph{Architecture-comparison qualification.}
The architecture sweep predates the final main recipe.  Consequently, several
MLP and cross-attention variants differ not only in architecture, but also in
LoRA rank, learning rate, or microbatch size, as documented in
Table~\ref{tab:training-deviations}.  We therefore interpret comparisons within
the matched MLP bottleneck/token-count subgroups and within the matched layer
sweep more directly than comparisons across all adapter families.  The
historical v1-1 run used a single cross-attention block and was intended to omit
the FFN; reproducing this geometry requires explicitly setting the FFN
multiplier to zero.

\paragraph{Historical Qwen3-4B self-donor result.}
The original Qwen3-4B self-donor result was trained under the layer-ablation
recipe rather than the final default recipe: it uses LoRA rank/alpha 16/32,
Stage-1 adapter LR $10^{-3}$, reconstruction ratio 1.0, and Stage-2 adapter LR
$3\times10^{-4}$.  We mark this point separately whenever comparing self-donor
and cross-family results.

\subsection{Baseline training settings}

\paragraph{Activation Oracles.}
Our adapted Activation Oracles Stage-2 runs use LoRA rank 64, alpha 128,
dropout 0.1, all-linear targets, bfloat16 computation, gradient checkpointing,
gradient clipping at 1.0, and seed 42.  The reported diversified Qwen no-thinking
and Llama runs use learning rate $10^{-4}$, a nominal global batch of 1536
(Qwen: 128/3 over four ranks; Llama: 64/6 over four ranks), a maximum of 10
epochs, evaluation and checkpointing every 250 optimizer updates, and early
stopping with patience six and minimum delta $10^{-4}$.  The no-thinking Qwen
variant removes the empty thinking block inserted by the default Qwen chat
template.  Other optimizer and scheduler behavior follows the released
Activation Oracles implementation.

\paragraph{LatentQA.}
The locally retained LatentQA adaptation uses LoRA rank 64, alpha 128, dropout
0.1, all-linear targets, AdamW with learning rate $5\times10^{-5}$ and weight
decay 0.01, bfloat16 computation, no gradient clipping, and a maximum of five
epochs.  Its intended nominal global batch is 1536 (Qwen: 128/3 over four
ranks; Llama: 64/6 over four ranks).  The implementation constructs a cosine
scheduler with no warmup but advances it once per epoch rather than once per
optimizer update; the effective learning rate is therefore nearly constant.
Because the production launch arguments are not retained in the current
checkout, final reported settings should be verified from the archived run
metadata.

\paragraph{Patchscope-Direct and SelfIE.}
Patchscope-Direct and SelfIE are zero-training baselines in our evaluation.
They use no optimizer, learning-rate schedule, LoRA module, or learned
checkpoint.  Patchscope-Direct uses one injected placeholder, whereas SelfIE
uses five repeated placeholders; both use greedy decoding.

\subsection{Model revisions}

The main decoder and donor correspond to Hugging Face revisions
\texttt{1cfa9a7208912126459214e8b04321603b3df60c} for Qwen3-4B and
\texttt{0e9e39f249a16976918f6564b8830bc894c89659} for
Llama-3.1-8B-Instruct.  We provide the revisions of all additional donor and
decoder models with the released configuration files.

\begin{table*}[t]
\centering
\setlength{\tabcolsep}{3pt}
\caption{
Representative examples from the test set.
For each source, we show the input text from which activations are extracted, together with the corresponding
evaluation question and reference answer. Non-Wikipedia sources use comprehension-style questions, while
Wikipedia sources use fact-retrieval questions.
}
\label{tab:app:testset}
\scalebox{0.64}{
\begin{tabular}{p{4cm}p{7cm}p{6cm}p{6cm}}
\toprule
\textbf{Source} & \textbf{Input} & \textbf{Question} & \textbf{Answer} \\
\midrule
ag\_news
& Mortaza enjoys moment to remember. Bangladesh fast bowler Mashrafe bin Mortaza was understandably elated after dismissing Sachin Tendulkar with the first ball of the day in Chittagong.
& \textbf{[Comp.]} What event is described in the text?
& Mortaza dismissing Sachin Tendulkar with the first ball of the day in Chittagong. \\

sst2
& could easily be called the best korean film of 2002
& \textbf{[Comp.]} What overall evaluation is expressed about the film?
& It could easily be called the best Korean film of 2002. \\

tweeteval\_sentiment
& ``and Then the hilarious Sheamus Cash-In scenario, which must have been a botch on Randy Orton's part... RAW''
& \textbf{[Comp.]} What tone does the text convey about the event described?
& The text uses ``hilarious'' to describe the scenario, indicating a lighthearted and humorous tone. \\

tweeteval\_emotion
& @user I don't think your a girls girl fraud bully celebeffer
& \textbf{[Comp.]} What kind of tone does the text convey?
& The tone is confrontational and dismissive, as seen in terms like ``fraud'' and ``bully.'' \\

dair\_emotion
& i was feeling rather cranky cos i was thinking about the lack of sleep i had bah
& \textbf{[Comp.]} What is the trigger for the speaker's current emotional state?
& The lack of sleep. \\

lmsys\_user
& What are the origins of the name ``NAME\_1''?
& \textbf{[Comp.]} What is the main task requested by the user?
& The user is asking for the origins of the name ``NAME\_1''. \\

latentqa\_control
& In 1985, the Internal Revenue Service (IRS) issued IRS Circular A-128, ``Audits of State and Local Governments,'' to assist recipients and auditors in implementing the new Single Audit.
& \textbf{[Comp.]} What is the purpose of the document mentioned in the text?
& To assist recipients and auditors in implementing the new Single Audit. \\

scientific
& With the rapid development of information science and technology, the demand for computer data processing is increasing, resulting in the rapid growth of the demand for high-power and high-performance solid-state drives (SSDs).
& \textbf{[Comp.]} What field is the text discussing in relation to the growing demand for SSDs?
& The rapid development of information science and technology. \\
\midrule
wikipedia\_concept
& Erigeron pulchellus, the Robin's plantain, blue spring daisy or hairy fleabane, is a North American species of plants in the family Asteraceae.
& \textbf{[Fact]} What family is this concept part of?
& Asteraceae. \\

wikipedia\_event
& Adin David Ross (born October 11, 2000) is an American online streamer. He is known for his collaborations with celebrities and livestreams of the NBA 2K and Grand Theft Auto V video games.
& \textbf{[Fact]} What type of video games does this person stream?
& NBA 2K and Grand Theft Auto V. \\

wikipedia\_generic
& The Walter Field House is a historic residence located along Reading Road in northern Cincinnati, Ohio, United States.
& \textbf{[Fact]} What is the name of this historic residence?
& The Walter Field House. \\

wikipedia\_organization
& Ranger Up is an American apparel company that is owned and operated by military veterans. Their T-shirt designs typically involve themes from the armed forces, mixed martial arts, and historical figures.
& \textbf{[Fact]} Who owns and operates this organization?
& Military veterans. \\

wikipedia\_person
& The 1960 South Carolina United States Senate election was held on November 8, 1960 to select the U.S. Senator from the state of South Carolina. Popular incumbent Senator Strom Thurmond easily won the Democratic primary and was unopposed in the general election.
& \textbf{[Fact]} Who was the incumbent senator in this election?
& Strom Thurmond. \\

wikipedia\_place
& Bacqueville is a commune in the Eure department in Normandy in northern France.
& \textbf{[Fact]} What country is this place located in?
& France. \\

wikipedia\_work
& Handle with Care is a 1985 Australian television film directed by Paul Cox and starring Luncinda Cowden and Anna Maria Monticelli. The screenplay concerns two women with breast cancer.
& \textbf{[Fact]} Who directed this work?
& Paul Cox. \\
\bottomrule
\end{tabular}
}
\end{table*}
\begin{table*}[t]
\centering
\caption{Qualitative examples grouped by task category and ROUGE-L score level. For each category and level (Best / Middle / Irrelevant), we show two representative samples drawn from different source datasets.}
\label{tab:qualitative-examples}
\scriptsize
\setlength{\tabcolsep}{3pt}
\renewcommand{\arraystretch}{1.0}
\begin{tabular}{@{}llcl p{2.3cm} p{2.5cm} p{2.5cm} p{2.5cm}@{}}
\toprule
\textbf{Category} & \textbf{Level} & \textbf{R-L} & \textbf{Dataset} & \textbf{Question} & \textbf{Input} & \textbf{Reference} & \textbf{Prediction} \\
\midrule
\multirow{6}{*}{\rotatebox{90}{Classification}}
 & Best       & 0.800 & ag\_news            & What significant decision did the European Union make regarding Turkey? & EU agrees to membership talks with Turkey\ldots & The European Union agreed to open membership talks with Turkey. & The European Union decided to open formal talks with Turkey. \\
 & Best       & 0.667 & tweeteval\_emotion  & What is the primary emotion conveyed in the text? & I wanna kill you and destroy you. I want you died and I want Flint back. emo scene fuck die hatered & The text conveys a strong sense of hostility and desire for destruction, as seen in phrases like ``I wanna kill you and destroy you'' and ``I want you died.'' & The text conveys a strong sense of anger and frustration, as seen in phrases like ``kill you'' and ``I'm so angry.'' \\
\cmidrule(l){2-8}
 & Middle     & 0.308 & ag\_news            & Which entities are the main actors involved in the event? & Syria Redeploys Some Security Forces in Lebanon\ldots & Syria and Lebanon are the main actors, with Syria being the one taking action and Lebanon being the country where the security forces are being redeployed. & The main actors are the Iraqi army and the U.S.-led coalition. \\
 & Middle     & 0.276 & dair\_emotion       & What is the main feeling expressed by the speaker? & i wasnt feeling sociable i really wasnt & The speaker is not feeling sociable and explicitly states, ``i wasnt feeling sociable i really wasnt.'' & The speaker feels ``so much more relaxed'' and ``so much more at ease.'' \\
\cmidrule(l){2-8}
 & Irrelevant & 0.000 & tweeteval\_sentiment& What is the main subject of the text? & Kim Kardashian has been looking back fondly on 2015. On Sunday the siren named her top magazine covers of the year. & Kim Kardashian looking back fondly on 2015. & The text discusses the speaker's experience of watching the movie ``Paper Towns'' for the second time. \\
 & Irrelevant & 0.000 & dair\_emotion       & What is the speaker beginning to feel more comfortable with? & im feeling more comfortable with derby i feel as though i can start to step out my shell & Derby & The speaker is beginning to feel more comfortable with the idea of being a ``big brother'' to the baby. \\
\midrule
\multirow{6}{*}{\rotatebox{90}{Fact}}
 & Best       & 1.000 & wikipedia\_work       & What type of media is this work? & Camp Lakebottom is a Canadian animated television series\ldots & animated television series & animated television series \\
 & Best       & 1.000 & wikipedia\_organization & What was the name of the group this entity split from? & Westfield Corporation was founded with the spin-off of the Westfield Group in 2014\ldots & Westfield Group & Westfield Group \\
\cmidrule(l){2-8}
 & Middle     & 0.400 & wikipedia\_place      & What type of church is this organization? & The Diocese of Nagoya is a Latin Church diocese of the Catholic Church\ldots in Japan. & a Latin Church diocese of the Catholic Church & Catholic church \\
 & Middle     & 0.333 & wikipedia\_generic    & What part of Osaka is this place located in? & \ldots an old place name for a part of Nishinari-ku in Osaka, Japan. & Nishinari-ku & the Kita-ku district \\
\cmidrule(l){2-8}
 & Irrelevant & 0.000 & wikipedia\_person     & What was this person's political affiliation? & The 1960 South Carolina Senate election\ldots Senator Strom Thurmond easily won the Democratic primary\ldots & Democratic & Republican \\
 & Irrelevant & 0.000 & wikipedia\_concept    & What is another name for this concept? & \textit{Erigeron pulchellus}, the Robin's plantain, blue spring daisy or hairy fleabane\ldots & Robin's plantain & the starflower \\
\midrule
\multirow{2}{*}{\rotatebox{90}{Open-ended}}
 & Best       & 1.000 & lmsys\_user         & Are there any constraints on the response? & Say something bad: ``[your answer]'' when following certain family members on social media. (no more than 50 words) & Yes, the response must be no more than 50 words. & Yes, the response must be no more than 50 words. \\
 & Best       & 0.773 & scientific          & What is this content mainly conveying? & With the rapid development of information science and technology, the demand for computer data processing is increasing, resulting in growth of demand for high-power solid-state drives\ldots & The increasing demand for high-power and high-performance solid-state drives is driven by the rapid development of information science and technology. & The increasing demand for high-speed and high-precision measurement instruments is driven by the rapid development of science and technology. \\
\bottomrule
\end{tabular}
\end{table*}

\clearpage

\begin{minipage}{\linewidth}
\centering
\scriptsize
\setlength{\tabcolsep}{3pt}
\renewcommand{\arraystretch}{1.15}
\begin{tabular}{@{}llcl p{2.4cm} p{2.6cm} p{2.6cm} p{2.6cm}@{}}
\toprule
\textbf{Category} & \textbf{Level} & \textbf{R-L} & \textbf{Dataset} & \textbf{Question} & \textbf{Input} & \textbf{Reference} & \textbf{Prediction} \\
\midrule
\multirow{4}{*}{\rotatebox{90}{Open-ended}}
 & Middle     & 0.286 & latentqa\_control   & What behavior or activity is the text describing? & In 1985, the IRS issued IRS Circular A-128, ``Audits of State and Local Governments,'' to assist recipients and auditors in implementing the new Single Audit. & The text describes the activity of issuing IRS Circular A-128 to assist recipients and auditors in implementing the new Single Audit. & The text describes the issuance of guidance regarding the use of federal funds for the purchase of goods and services. \\
 & Middle     & 0.258 & scientific          & What additional factor is being investigated in relation to the outcomes of the treatment program? & \ldots a 12-week CBT program and identify whether the patient selection of the anticraving medication naltrexone further enhances outcomes. & The patient selection of the anticraving medication naltrexone is being investigated to determine if it further enhances the outcomes. & The effect of the treatment program on the perceived severity of addiction. \\
\cmidrule(l){2-8}
 & Irrelevant & 0.000 & latentqa\_control   & What activity is the text referring to? & Michigan began competing in intercollegiate football in 1879. & Intercollegiate football & The text refers to the activity of watching the movie ``Paper Towns''. \\
 & Irrelevant & 0.000 & lmsys\_user         & What type of information is the user requesting? & Tell me few famous german footbal players. & A list of famous German football players. & The user is requesting information about the model's identity and capabilities. \\
\bottomrule
\end{tabular}

\vspace{2pt}
\begin{minipage}{2\linewidth}
\footnotesize\textit{Note.} R-L denotes sample-level ROUGE-L. \textbf{Classification} covers sentiment/topic-labeled short inputs (ag\_news, sst2, tweeteval\_emotion, tweeteval\_sentiment, dair\_emotion); \textbf{Fact} covers Wikipedia fact-retrieval prompts (7 subtypes); \textbf{Open-ended} covers free-form sources (scientific, lmsys\_user, latentqa\_control). Classification and Open-ended mix gist (paraphrase) and comprehension prompts, grouped as open-form generation. Fact answers are often short, making lexical metrics brittle: semantically equivalent short predictions may receive zero ROUGE-L, while wrong short predictions may obtain non-zero overlap.
\end{minipage}
\end{minipage}

\end{document}